\documentclass{clv2}
 
\usepackage{url}
\usepackage{xspace}
\usepackage[normalem]{ulem}
\usepackage[leftmargin=4mm,rightmargin=1mm]{quoting}
\usepackage{amsmath}

\newcommand{\code}[1]{{\small \texttt{#1}}} 

\newcommand{\owl}{\textsc{owl}\xspace}

\newcommand{\nlowl}{Natural\textsc{owl}\xspace}

\newcommand{\rdf}{\textsc{rdf}\xspace}
\newcommand{\rdfs}{\textsc{rdf schema}\xspace}

\newcommand{\nlg}{\textsc{nlg}\xspace}
\newcommand{\nl}{\textsc{nl}\xspace}

\newcommand{\protege}{Prot\'eg\'e\xspace}
\newcommand{\etalt}{et al.\xspace}
\newcommand{\pos}{\textsc{pos}\xspace}

\newcommand{\ilp}{\textsc{ilp}\xspace}
\newcommand{\ilpnlg}{\textsc{ilpnlg}\xspace}
\newcommand{\ilpnlgextend}{\textsc{ilpnlgextend}\xspace}
\newcommand{\ilpnlgapprox}{\textsc{ilpnlgapprox}\xspace}
\newcommand{\pipeline}{\textsc{pipeline}\xspace}
\newcommand{\pipelinestoch}{\textsc{pipelinestoch}\xspace}
\newcommand{\pipelineshort}{\textsc{pipelineshort}\xspace}
\newcommand{\pipelinebeam}{\textsc{pipelinebeam}\xspace}
\newcommand{\pipelineshortnln}{\textsc{pipelineshort*}\xspace}

\runningtitle{Generating Texts with Integer Linear Programming}

\runningauthor{Lampouras \& Androutsopoulos}

\begin{document}
\title{Producing Compact Texts with Integer Linear Programming in Concept-to-Text Generation}

\author{Gerasimos Lampouras\thanks{
Department of Informatics, Athens University of Economics and Business, Patission 76, 104 34 Athens, Greece.
E-mail: lampouras06@aueb.gr}}
\affil{Department of Informatics, Athens University of Economics and Business}

\author{Ion Androutsopoulos\thanks{Department of Informatics, Athens University of Economics and Business, Patission 76, 104 34 Athens, Greece.
E-mail: ion@aueb.gr}}
\affil{Department of Informatics, Athens University of Economics and Business}

\maketitle

\begin{abstract}
Concept-to-text generation typically employs a pipeline architecture, which often leads to suboptimal texts. Content selection, for example, may greedily select the most important  facts, which may require, however, too many words to express, and this may be undesirable when space is limited or expensive. Selecting other facts, possibly only slightly less important, may allow the lexicalization stage to use much fewer words, or to report more facts in the same space. Decisions made during content selection and lexicalization may also lead to more or fewer sentence aggregation opportunities, affecting the length and readability of the resulting texts. Building upon on a publicly available state of the art natural language generator for Semantic Web ontologies, this article presents an Integer Linear Programming model that, unlike pipeline architectures, jointly considers choices available in content selection, lexicalization, and sentence aggregation to avoid greedy local decisions and produce more compact texts, i.e., texts that report more facts per word. Compact texts are desirable, for example, when generating advertisements to be included in Web search results, or when summarizing structured information in limited space. An extended version of the proposed model also considers a limited form of referring expression generation and avoids redundant sentences. An approximation of the two models can be used when longer texts need to be generated. Experiments with three ontologies confirm that the proposed models lead to more compact texts, compared to pipeline systems, with no deterioration or with improvements in the perceived quality of the generated texts. 
\end{abstract}

\section{Introduction} \label{introduction}

The Semantic Web \cite{BernersLee2001,Shadbolt2006} and the growing popularity of Linked Data (data that are published using Semantic Web technologies) have renewed interest in concept-to-text natural language generation (\nlg), especially text generation from ontologies \cite{Bontcheva2005,Mellish2006,Galanis2007,Mellish2008,Schwitter2008,Schwitter2010,Liang2011b,Williams2011,Androutsopoulos2013}. An ontology provides a conceptualization of a knowledge domain (e.g., consumer electronics, diseases) by defining the classes and subclasses of the individuals (entities) in the domain, the possible relations between them etc. The current standard to specify Semantic Web ontologies is \owl \cite{Horrocks2003,Grau2008}, a formal language based on description logics \cite{Baader2002}, \rdf, and \rdfs \cite{Antoniou2008}.\footnote{Most Linked Data currently use only \rdf and \rdfs, but \owl is in effect a superset of \rdfs and, hence, methods to produce texts from \owl also apply to Linked Data. Consult also \url{http://linkeddata.org/}.}  Given an \owl ontology for a knowledge domain, one can publish on the Web machine-readable statements about the domain (e.g., available products, known diseases, their features or symptoms), with the statements having formally defined semantics based on the ontology. \nlg can then produce texts describing classes or individuals of the ontology (e.g., product descriptions, information about diseases) from the same statements.\footnote{Following common practice in Semantic Web research, we often use the term `ontology' to refer jointly to terminological knowledge (TBox statements) that establishes a conceptualization of a knowledge domain, and assertional knowledge (ABox statements) that describes particular individuals.} This way the same information becomes more easily accessible to both computers (which read the machine-readable statements) and end-users (who read the texts), which is one of the main goals of the Semantic Web.

\nlg systems typically employ a pipeline architecture \cite{ReiterDale2000}. Firstly, content selection chooses the logical facts (axioms, in the case of an \owl ontology) to be expressed in the text to be generated. The purpose of the next stage, text planning, ranges from simply ordering the facts to be expressed, in effect also ordering the sentences that will express them, to making more complex decisions about the rhetorical structure of the text. Lexicalization then selects the words and syntactic structures to express each fact as a single sentence. Sentence aggregation may then combine shorter sentences into longer ones. Another component generates appropriate referring expressions (pronouns, noun phrases etc.), and surface realization produces the final text, based on internal representations of the previous decisions. Each stage of the pipeline in effect performs a local optimization, constrained by decisions of the previous stages, and largely unaware of the consequences of its own decisions on the subsequent stages. 

The pipeline architecture has engineering advantages (e.g., it is easier to specify and monitor the input and output of each stage), but produces texts that may be suboptimal, since the decisions of the generation stages are actually co-dependent \cite{Danlos1984,Marciniak2005b,Belz2008}. 
Content selection, for example, may greedily select the most important facts among those that are relevant to the purpose of the text, but these facts may require too many words to express, which may be undesirable when space is limited or expensive. Selecting other facts, possibly only slightly less important, may allow the lexicalization stage to use much fewer words, or to report more facts in the same space. Decisions made during content selection and lexicalization (facts to express, words and syntactic structures to use) may also lead to more or fewer sentence aggregation opportunities, affecting the length and readability of the texts. Some of these issues can be addressed by over-generating at each stage (e.g., producing several alternative sets of facts at the end of content selection, several alternative lexicalizations etc.) and employing a final ranking component to select the best combination \cite{Walker2001}. This over-generate and rank approach, however, may also fail to find an optimal solution, and generates an exponentially large number of candidate solutions when several components are pipelined. 

In this article, we present an Integer Linear Programming (\ilp) model that, unlike pipeline architectures, jointly considers choices available in content selection, lexicalization, and sentence aggregation to avoid greedy local decisions and produce more compact texts, i.e., texts that report more facts per word. Compact texts are desirable, for example, when generating short product descriptions to be included as advertisements in Web search results \cite{Thomaidou2013,Thomaidou2014}.\footnote{See also \url{http://www.google.com/ads/}.} Question answering may also involve generating a natural language summary of facts (e.g., \rdf triples) related to a question, without exceeding a maximum text length \cite{Tsatsaronis2012BioASQ}; the more compact the summary, the more facts can be reported in the available space, increasing the chances of reporting the information sought by the user.\footnote{Consult also \url{http://nlp.uned.es/clef-qa/} and \url{http://www.bioasq.org/}.} 
Compact texts are also desirable when showing texts on devices with small screens \cite{CorstonOliver2001} or as subtitles \cite{Vandeghinste2004}.\footnote{See also the smartphone application Acropolis Rock (\url{http://acropolisrock.com/}), which uses \nlowl to describe historical monuments; a video is available at \url{https://www.youtube.com/watch?v=XMzdTir6Gas}. The subtitles of the virtual museum guide of Galanis \etalt \shortcite{Galanis2009} are also generated by \nlowl; see the video at \url{http://vimeo.com/801099}.} 

If an importance score is available for each fact, our model can take it into account to maximize the total importance (instead of the total number) of the expressed facts per word. The model itself, however, does not produce importance scores; we assume that the scores are produced by a separate process \cite{Barzilay2005,Demir2010}, not included in our content selection. For simplicity, in the experiments of this article we treat all the facts as equally important. An extended version of our \ilp model also considers a limited form of referring expression generation, where the best name must be chosen per individual or class among multiple alternatives. The extended model also avoids sentences that report information that is obvious (to humans) from the names of the individuals and classes (e.g., ``A red wine is a kind of wine with red color'').  Experiments with three \owl ontologies from very different knowledge domains (wines, consumer electronics, diseases) confirm that our models lead to more compact texts, compared to pipeline systems with the same components, with no deterioration or with improvements in the perceived quality of the generated texts. Although solving \ilp problems is in general \textsc{np}-hard \cite{Karp72}, off-the-shelf \ilp solvers can be used. The available solvers guarantee finding a globally optimum solution, and they are very fast in practice in the \ilp problems we consider, when the the number of available facts (per individual or class being described) is small. We also present an approximation of our \ilp models, which is more efficient when the number of available facts is larger and longer texts need to be generated.

Our \ilp models (and approximations) have been embedded in \nlowl \cite{Androutsopoulos2013}, an \nlg system for \owl, as alternatives to the system's original pipeline architecture. We base our work on \nlowl, because it is the only open-source \nlg system for \owl that implements all the processing stages of a typical \nlg system \cite{ReiterDale2000}, it is extensively documented, and has been tested with several ontologies.  The processing stages and linguistic resources of \nlowl are typical of \nlg systems \cite{Mellish2006a}. Hence, we believe that our work is, at least in principle, also applicable to other \nlg systems. Our \ilp models do not directly consider text planning, but rely on the (external to the \ilp model) text planner of \nlowl. We hope to include more text planning and referring expression generation decisions directly in our \ilp model in future work. We also do not consider surface realization, since it is not particularly interesting in \nlowl; all the decisions have in effect already been made by the time this stage is reached.

The remainder of this article is structured as follows. Section~\ref{NaturalOWL} below provides background information about \nlowl. Section \ref{ILPModel} defines our \ilp models. Section~\ref{Complexity} discusses the computational complexity of our \ilp models, along with the more efficient approximation that can be used when then number of available facts is large. Section~\ref{Experiments} presents our experiments. Section \ref{RelatedWork} discusses previous related work. Section \ref{Conclusions} concludes and proposes future work.

\section{Background Information about NaturalOWL} \label{NaturalOWL}

\nlowl produces texts describing classes or individuals (entities) of an \owl ontology (e.g., descriptions of types of products or particular products). Given an \owl ontology and a particular target class or individual to describe, \nlowl first scans the ontology for \owl statements relevant to the target. If the target is the class \code{StEmilion}, for example, a relevant \owl statement may be the following.\footnote{This example is based on the Wine Ontology, one of the ontologies of our experiments (see Section~\ref{Experiments}).}
{\small
\begin{verbatim}
   SubclassOf(:StEmilion 
     ObjectIntersectionOf(:Bordeaux                     
       ObjectHasValue(:locatedIn :stEmilionRegion)
       ObjectHasValue(:hasColor :red)
       ObjectHasValue(:hasFlavor :strong)          
       ObjectHasValue(:madeFrom :cabernetSauvignonGrape) 
       ObjectMaxCardinality(1 :madeFrom)))
\end{verbatim}
}
\noindent The statement above defines \code{StEmilion} as the intersection of: (i) the class of \code{Bordeaux} wines; (ii) the class of all individuals whose \code{locatedIn} property has (for each individual) \code{stEmilionRegion} among its values (\owl properties are generally many-valued); (iii)--(v) the classes of individuals whose \code{hasColor}, \code{hasFlavor}, and \code{madeFromGrape} property values include  \code{red}, \code{strong}, and \code{cabernetSauvignonGrape}, respectively, without excluding wines that have additional values in these properties; and (vi) the class of individuals whose \code{madeFromGrape} property has exactly one value; hence, a St.\ Emilion wine is made \emph{exclusively} from Cabernet Sauvignon grapes. 

\nlowl then converts each relevant statement into (possibly multiple) \emph{message triples} of the form $\left<S, R, O\right>$, where $S$ is an individual or class, $O$ is another individual, class, or datatype value, and $R$ is a relation (property) that connects $S$ to $O$.\footnote{For simplicity, we omit some details about message triples. Consult Androutsopoulos \etalt  \shortcite{Androutsopoulos2013} for more information about message triples and their relation to \rdf triples.} For example, the \code{ObjectHasValue(:madeFrom :cabernetSauvignonGrape)} part of the \owl statement above is converted to the message triple $<$\code{:StEmilion}, \code{:madeFrom}, \code{:cabernetSauvignonGrape}$>$. Message triples are similar to \rdf triples, but they are easier to express as sentences. Unlike \rdf triples, the relations ($R$) of the message triples may include \emph{relation modifiers}. For example, the \code{ObjectMaxCardinality(1 :madeFrom)} part of the \owl statement above is turned into the message triple $<$\code{:StEmilion}, \code{maxCardinality(:madeFrom)}, 1$>$, where \code{maxCardinality} is a relation modifier. 
Message triples may also contain conjunctions or disjunctions as their $O$, as in $<$\code{:ColoradoTickFever}, \code{:hasSymptom}, \code{and(:fatigue, :headache, :myalgia)}$>$.\footnote{This example is from the Disease Ontology, another ontology used in our experiments.} We use the terms `fact' and `message triple' as synonyms in the remainder of this article. 

Having produced the message triples, \nlowl consults a user model to select the most 
important ones,
and orders the selected triples according to manually authored text plans. Later processing stages convert each message triple to an abstract  sentence representation, aggregate sentences to produce longer ones, and produce appropriate referring expressions (e.g., pronouns). The latter three stages require a \emph{sentence plan} for each relation ($R$), while the last stage also requires \emph{natural language names} (\nl names) for the individuals and classes of the ontology. Rougly speaking, a sentence plan specifies how to generate a sentence to express a message triple involving a particular relation ($R$), whereas an \nl name specifies how to generate a noun phrase to refer to a class or individual by name. We provide more information about sentence plans and \nl names in the following subsections. If sentence plans and \nl names are not supplied, \nlowl automatically produces them by tokenizing the \owl identifiers of the relations, individuals, and classes of the ontology, acting as a simple \emph{ontology verbalizer} \cite{Cregan2007,Kaljurand2007,Schwitter2008,HalaschekWiener2008,Schutte2009,Power2010b,Power2010,Schwitter2010,Stevens2011,Liang2011b}.  The resulting texts, however, are of much lower quality \cite{Androutsopoulos2013}. For example, the resulting text from the \owl statement above would be:

\begin{quoting}
{\small
\noindent St Emilion is Bordeaux. St Emilion located in St Emilion Region. St Emilion has color Red. St Emilion has flavor Strong. St Emilion made from grape exactly 1: Cabernet Sauvignon Grape. 
}
\end{quoting}

\noindent By contrast, when appropriate sentence plans and \nl names are provided, \nlowl produces the following text:

\begin{quoting}
{\small 
\noindent St.\ Emilion is a kind of red, strong Bordeaux from the St.\ Emilion region. It is made from exactly one grape variety: Cabernet Sauvignon grapes.
}
\end{quoting}

In this article, we assume that appropriate sentence plans and \nl names are supplied for each ontology. They can be manually constructed  using a \protege plug-in that accompanies \nlowl \cite{Androutsopoulos2013}.\footnote{Consult \url{http://protege.stanford.edu/}. \nlowl and its \protege plug-in are available from \url{http://nlp.cs.aueb.gr/software.html}.} Semi-automatic methods can also be used to extract and rank candidate sentence plans and \nl names from the Web, with a human selecting the best among the most highly ranked ones; in this case, it has been shown that high quality sentence plans and \nl names can be constructed in a matter of a few hours (at most) per ontology \cite{Lampouras2015}.

\subsection{The Natural Language Names of NaturalOWL} \label{NLnamesBackground}

In \nlowl, an \nl name is a sequence of slots. The contents of the slots are concatenated to produce a noun phrase that names a class or individual. Each slot is accompanied by annotations specifying how to fill it in; the annotations may also provide linguistic information about the contents of the slot. For example, we may specify that the English \nl name of the class \code{:TraditionalWinePiemonte} is the following.\footnote{The \nl names and sentence plans of \nlowl are actually represented in \owl, as instances of an ontology that describes the domain-dependent linguistic resources of the system.}

\begin{center}
[\,]$^{1}_{\textit{article}, \, \textit{indef}, \, \textit{agr}=3}$ 
[traditional]$^{2}_{\textit{adj}}$ 
[wine]$^{3}_{\textit{headnoun}, \, \textit{sing}, \, \textit{neut}}$
[from]$^{4}_{\textit{prep}}$ 
[\,]$^{5}_{\textit{article}, \, \textit{def}}$ 
[Piemonte]$^{6}_{\textit{noun}, \, \textit{sing}, \, \textit{neut}}$ 
[region]$^{7}_{\textit{noun}, \, \textit{sing}, \, \textit{neut}}$ 
\end{center}

\noindent The first slot is to be filled in with an indefinite article, whose number should agree with the third slot.  The second slot is to be filled in with the adjective `traditional'. The third slot with the neuter noun `wine', which will also be the head (central) noun of the noun phrase, in singular number, and similarly for the other slots. \nlowl makes no distinctions between common and proper nouns, but it can be instructed to capitalize particular nouns (e.g., `Piemonte'). In the case of the message triple  $<$\code{:wine32}, \code{instanceOf}, \code{:TraditionalWinePiemonte}$>$, the \nl name above would allow a sentence like ``This is \emph{a traditional wine from the Piemonte region}'' to be produced. 

The slot annotations allow \nlowl to automatically adjust the \nl names. For example, the system also generates comparisons to previously encountered individuals or classes, as in ``Unlike the previous products that you have seen, which were all \emph{traditional wines from the Piemonte region}, this is a French wine''. In this particular example, the head noun (`wine') had to be turned into plural. Due to number agreement, its article also had to be turned into plural; in English, the plural indefinite article is void, hence the article of the head noun was omitted. As a further example, we may specify that the \nl name of the class \code{FamousWine} is the following.

\begin{center}
[\,]$^{1}_{\textit{article}, \, \textit{indef}, \, \textit{agr}=3}$ 
[famous]$^{2}_{\textit{adj}}$ 
[wine]$^{3}_{\textit{headnoun}, \, \textit{sing}, \, \textit{neut}}$
\end{center}

\noindent 
If the triples $<$\code{:wine32}, \code{instanceOf}, \code{:TraditionalWinePiemonte}$>$ and $<$\code{:wine32}, \code{instanceOf}, \code{:FamousWine}$>$ were to be expressed, \nlowl would then produce the single, aggregated sentence ``This is a famous traditional wine from the Piemonte region'', instead of two separate sentences ``This is a traditional wine from the Piemonte region'' and ``It is a famous wine''. The annotations of the slots, which indicate for example which words are adjectives and head nouns, are used by the sentence aggregation component to appropriately combine the two sentences. The referring expression generation component also uses the slot annotations to identify the gender of the head noun, when a pronoun has to be generated (e.g., `it' when the head noun is neuter). 

We can now define more precisely \nl names. An \nl name is a sequence of one or more slots. Each slot is accompanied by annotations requiring it to be filled in with exactly one of the following:\footnote{\nlowl also supports Greek. The possible annotations for Greek \nl names (and sentence plans, see below) are slightly different, but in this article we consider only English \nl names (and sentence plans).}

\smallskip
(1) \textit{An article}, definite or indefinite, possibly to agree with a noun slot. 

(2) \textit{A noun flagged as the head}. The number of the head noun must also be specified. 

(3) \textit{An adjective flagged as the head}. For example, the \nl name of the individual \code{:red} may consist of a single slot, to be filled in with the adjective `red', which will also be the head of the \nl name. The number and gender of the head adjective must be specified. 

(4) \textit{Any other noun or adjective}, (5) \textit{a preposition}, or (6) \textit{any fixed (canned) string}. 

\smallskip
\noindent Exactly one head (noun or adjective) must be specified per \nl name.  For nouns and adjectives, the \nl name may require a particular inflectional form to be used (e.g., in a particular number, case, or gender), or it may require an inflectional form that agrees with another noun or adjective slot.\footnote{We use \textsc{simplenlg} \cite{Gatt2009} to generate the inflectional forms of nouns, adjectives, verbs.} Multiple \nl names can also be provided for the same individual or class, to produce more varied texts.  

When providing \nl names, an individual or class can also be declared to be \emph{anonymous}, indicating that \nlowl should avoid referring to it by name. For example, in a museum ontology, there may be a 
particular coin whose \owl identifier is \code{:exhibit49}. We may not wish to provide an \nl name for this individual (it may not have an English name); and we may want \nlowl to avoid referring to the coin by tokenizing its identifier (``exhibit 49''). By declaring the coin as anonymous, \nlowl would use only the \nl name of its class (e.g., ``this coin''), simply ``this'', or a pronoun.

\subsection{The Sentence Plans of NaturalOWL} \label{sentencePlansBackground}

In \nlowl, a sentence plan for a relation $R$ specifies how to construct a sentence to express any message triple of the form $\left<S, R, O\right>$. Like \nl names, sentence plans are sequences of slots with annotations specifying how to fill the slots in. The contents of the slots are concatenated to produce the sentence. For example, the following is a sentence plan for the relation \code{:madeFrom}.
\begin{center}
[$\mathit{ref}(S)$]$^{1}_{\textit{nom}}$
[make]$^{2}_{\textit{verb}, \, \textit{passive}, \, \textit{present}, \, \textit{agr}=1, \, \textit{polarity}=+}$ 
[from]$^{3}_{prep}$ 
[$\mathit{ref}(O)$]$^{4}_{\textit{acc}}$
\end{center}
Given the message triple $<$\code{:StEmilion}, \code{:madeFrom}, \code{:cabernetSauvignonGrape}$>$, the sentence plan would lead to sentences like ``St.\ Emilion is made from Cabernet Sauvignon grapes'', or ``It is made from  Cabernet Sauvignon grapes'', assuming that appropriate \nl names have been provided for \code{:StEmilion} and \code{:cabernetSauvignonGrape}. Similarly, given  $<$\code{:Wine}, \code{:madeFrom}, \code{:Grape}$>$, the sentence plan above would lead to sentences like ``Wines are made from grapes'' or ``They are made from grapes'', assuming again appropriate \nl names. As another example, the following sentence plan can be used with the relations \code{:hasColor} and \code{:hasFlavor}. 
\begin{center}
[$\mathit{ref}(S)$]$^{1}_{\textit{nom}}$
[be]$^{2}_{\textit{verb}, \, \textit{active}, \, \textit{present}, \, \textit{agr}=1, \, \textit{polarity}=+}$
[$\mathit{ref}(O)$]$^{3}_{\textit{nom}}$
\end{center}
For the message triples $<$\code{:StEmilion}, \code{:hasColor}, \code{:red}$>$ and $<$\code{:StEmilion}, \code{:hasFlavor}, \code{:strong}$>$, it would produce the sentences ``St.\ Emilion is red'' and ``St.\ Emilion is strong'', respectively. 
 
The first sentence plan above, for \code{:madeFrom}, has four slots. The first slot is to be filled in with an automatically generated referring expression (e.g., pronoun or name) for $S$, in nominative case. The verb of the second slot is to be realized in passive voice, present tense, and positive polarity (as opposed to expressing negation) and should agree (in number and person) with the referring expression of the first slot ($\textit{agr}=1$). The third slot is filled in with the preposition `from', and the fourth slot with an automatically generated referring expression for $O$, in accusative case. 

\nlowl has built-in sentence plans for domain-independent relations (e.g., \code{isA}, \code{instanceOf}). For example, $<$\code{:StEmilion}, \code{isA}, \code{:Bordeaux}$>$ is expressed as ``St.\ Emilion is a kind of Bordeaux'' using the following built-in sentence plan; the last slot requires the \nl name of $O$ without article.
\begin{center}
[$\mathit{ref}(S)$]$^1_{\textit{nom}}$
[be]$^2_{\textit{verb}, \, \textit{active}, \, \textit{present}, \, \textit{agr}=1, \, \textit{polarity}=+}$
[``a kind of'']$^3_{\textit{string}}$
[$\mathit{name}(O)$]$^4_{\textit{noarticle}, \textit{nom}}$
\end{center}

Notice that the sentence plans are not simply slotted string templates (e.g., ``$X$ is made from $Y$''). Their linguistic annotations (e.g., \pos tags, agreement, voice, tense, cases) along with the annotations of the \nl names allow \nlowl to produce more natural sentences (e.g., turn the verb into plural when the subject is in plural), produce appropriate referring expressions (e.g., pronouns in the correct cases and genders), and aggregate shorter sentences into longer ones. For example, the annotations of the \nl names and sentence plans allow \nlowl to produce the aggregated sentence ``St.\ Emilion is a kind of red Bordeaux made from Cabernet Sauvignon grapes'' from the  triples $<$\code{:StEmilion}, \code{isA}, \code{:Bordeaux}$>$, $<$\code{:StEmilion}, \code{:hasColor}, \code{:red}$>$, $<$\code{:StEmilion}, \code{:madeFrom}, \code{:cabernetSauvignonGrape}$>$, instead of three sentences. 

We can now define more precisely sentence plans. A sentence plan is a sequence of slots. Each slot is accompanied by annotations requiring it to be filled in with exactly one of the following:\footnote{For simplicity, we omit some details and  functionality of sentence plans that are not relevant to the work of this article. More details can be found elsewhere \cite{Androutsopoulos2013}.}

\smallskip
(1) \textit{A referring expression for the $S$} (a.k.a.\  the \emph{owner}) of the triple, in a particular case.

(2) \textit{A verb} in a particular polarity and inflectional form (e.g., tense, voice), possibly to agree with another slot. 

(3) \textit{A noun or adjective} in a particular form, possibly to agree with another slot.

(4) \textit{A preposition}, or (5) \textit{a fixed string}. 

(6) \textit{A referring expression for the $O$} (a.k.a.\ the \emph{filler}) of the triple, in a particular case.
\smallskip

Multiple sentence plans can be provided per relation, to produce more varied texts and  increase sentence aggregation opportunities. Sentence plans for message triples that involve relation modifiers (e.g., $<$\code{:StEmilion}, \code{maxCardinality(:madeFrom)}, 1$>$) are automatically produced from the sentence plans for the corresponding relations without modifiers (e.g., $<$\code{:StEmilion}, \code{:madeFrom}, \code{:cabernetSauvignonGrape}$>$).

\subsection{Importance Scores} \label{InterestReasoning}

Some message triples can lead to sentences that sound redundant, because they report relations that are obvious (to humans) from the \nl names of the individuals or classes, as in the sentence ``A red wine is a kind of wine with red color''. The sentence of our example reports the following two message triples: 
\begin{center}
$<$\code{:RedWine}, \code{isA}, \code{:Wine}$>$, 
$<$\code{:RedWine}, \code{:hasColor}, \code{:Red}$>$\\
\end{center}

\noindent Expressed separately, the two triples would lead to the sentences ``A red wine is a kind of wine'' and ``A red wine has red color'', but \nlowl aggregates them into a single sentence. It is obvious that a red wine is a wine with red color and, hence, the two triples above should not be expressed. Similarly, the following triple leads to the sentence ``A white Bordeaux wine is a kind of Bordeaux'', which again seems redundant.
\begin{center}
$<$\code{:WhiteBordeaux}, \code{isA}, \code{:Bordeaux}$>$
\end{center}

\nlowl allows message triples to be assigned \emph{importance scores} indicating how important (or interesting) it is to convey each message triple to different user types or particular users. Assigning a zero importance score to a message triple instructs \nlowl to avoid expressing it. The importance scores can be constructed manually or by invoking an external user modeling component \cite{Androutsopoulos2013}. An additional mechanism of \nlowl assigns zero importance scores to message triples like the ones above, which report relations that are obvious from the \nl names; this is achieved by using heuristics discussed elsewhere \cite{Lampouras2015}. In the experiments of this article, we use the zero importance scores that \nlowl automatically assigns to some message triples, but we treat all the other message triples as equally important for simplicity.

\section{Our Integer Linear Programming Models}\label{ILPModel}

\begin{figure}
\center
\includegraphics[width=\columnwidth]{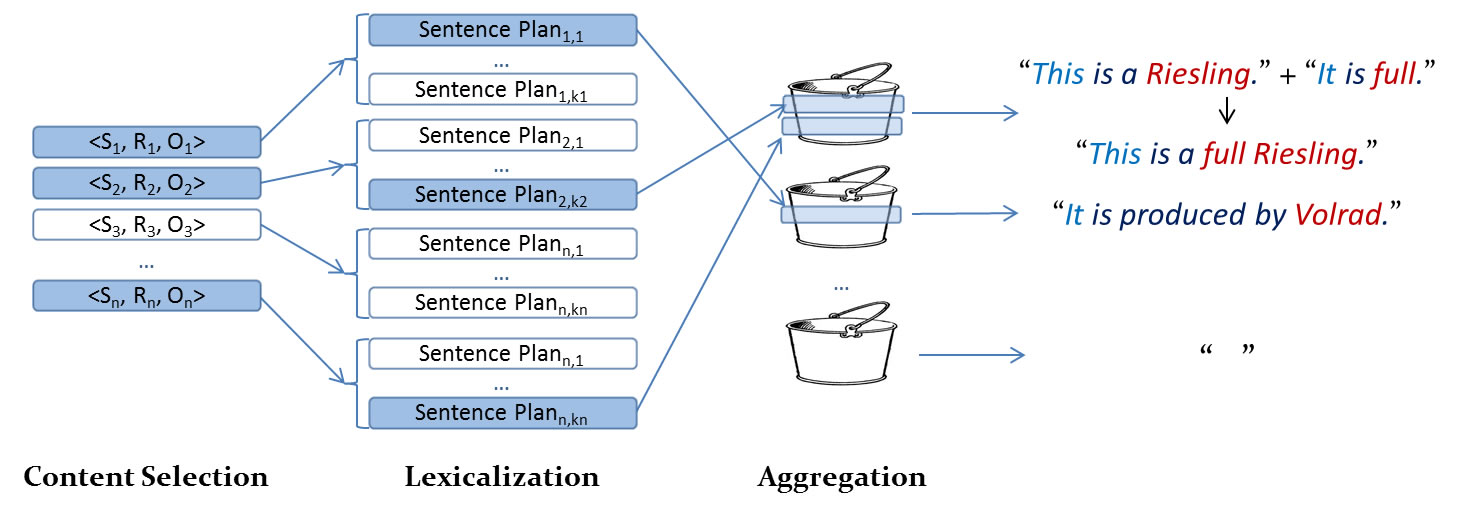}
\caption{Illustration of the main decisions of our first \ilp model.}
\label{ILPModelChoices}
\end{figure}

We now discuss our Integer Linear Programming (\ilp) models, starting from the first, simpler version, which considers choices available in content selection, lexicalization, and sentence aggregation. Figure~\ref{ILPModelChoices} illustrates the main decisions of the model. For content selection, the model decides which of the available facts (message triples) should be expressed. For lexicalization, it decides which sentence plan should be used for each fact that will be expressed, assuming that multiple sentence plans are available per fact. For sentence aggregation, it decides which simple sentences (each reporting a single fact) should be aggregated to form longer sentences, by partitioning the simple sentences (or equivalently the message triples they express) into groups (shown as buckets in Fig.~\ref{ILPModelChoices}). After using the \ilp model, the aggregation rules of \nlowl \cite{Androutsopoulos2013} are applied separately to the simple sentences of each group (bucket) to obtain a single aggregated sentence per group.\footnote{The sentences of each group can always be aggregated, since they describe the same individual or class. If no better aggregation rule applies, a conjunction of the sentences in the group can be formed.} To keep the \ilp model simpler, the model itself does not control which particular aggregation rules will be applied to each group. The number of groups (buckets) is fixed, equal to the maximum number of (aggregated) sentences that the model can generate per text. To avoid generating very long aggregated sentences, the number of simple sentences that can be placed in each group (bucket) cannot exceed a fixed upper limit (the same for all groups). Groups left empty produce no sentences. 

Our second, extended \ilp model is very similar, but also performs a limited form of referring expression generation by selecting among multiple alternative \nl names; it also takes into account that using a particular \nl name may make expressing some other facts redundant (Section~\ref{InterestReasoning}). By contrast, the first, simpler \ilp model assumes that a single \nl name is available per individual and class (hence, no choice of \nl names is needed) and does not try to avoid expressing redundant facts. In both models, a single (selected, or the only available one) \nl name is picked per individual or class (unless the individual or class is marked as anonymous, see Section~\ref{NLnamesBackground}), and it is used throughout the particular text being generated. Neither of the two models considers other referring expression generation decisions (e.g., whether to use a pronoun or a demonstrative noun phrase like ``this wine'', as opposed to repeating the \nl name of a wine). The existing referring expression generation component of \nlowl \cite{Androutsopoulos2013} is subsequently invoked (after using the \ilp models) to decide if the picked \nl name, a pronoun, or a demonstrative noun phrase should be used wherever a reference to an individual or class is needed in the text being generated.

A further limitation of our models is that they do not directly consider text planning, relying on the (external to the \ilp models) text planner of \nlowl instead. The text planner is invoked (before using the \ilp models) to partition the available message triples (the triples about the individual or class to be described) into \emph{topical sections}; for example, message triples about the size, weight, and material of an electronic product may be placed in one section, and triples about the functions and features of the product in another one. This step is needed, because our \ilp models never aggregate together sentences expressing facts from different topical sections, to avoid producing aggregated sentences that sound unnatural. The text planner is also invoked after using one of the \ilp models, to order each group of simple sentences that the model has decided to aggregate. As already noted, each aggregated sentence is produced by applying the aggregation rules of \nlowl to a group (bucket) of simple sentences, but the rules presuppose that the simple sentences to be aggregated are already ordered, which is why the text planer is invoked at this point. After applying the aggregation rules to each group of (ordered) simple sentences, the text planner is also used to order the topical sections, and the (now aggregated) sentences within each section.

\subsection{Our First ILP Model} \label{OurModel} 

Let us now focus on our first \ilp model. As already noted, this model assumes that there is a single \nl name per individual and class (excluding anonymous ones). Furthermore, the model assumes that all the \nl names are short and  approximately equally long.  

Let $F = \{f_1, \dots, f_n\}$ be the set of all the available facts $f_{i}$ about the target individual or class $S$ to be described. Recall that we use the term `fact' as a synonym of `message triple'. For each fact $f_i = \left<S, R_i, O_i\right>$, we assume that a set $P_{i} = \{p_{i1}, p_{i2}, \dots\}$ of alternative sentence plans is available; facts with the same relation ($R_i$) have the same set of sentence plans ($P_i$). Recall, also, that each sentence plan $p_{ik}$ specifies how to express $f_i$ as an alternative single sentence, and that a sentence plan is a sequence of slots, along with instructions specifying how to fill the slots in. 

We call \emph{elements} the unordered slots of a sentence plan along with their instructions, but with $S_i$ and $O_i$ accompanied by the individuals, classes, or datatype values they refer to. In the first example of Section~\ref{sentencePlansBackground}, there are four elements:
[$\mathit{ref}(S$ = \code{:StEmilion}$)$], [make]$_{\textit{present}, \; \textit{passive}}$, [from], [$\mathit{ref}(O$ = \code{:cabernetSauvignonGrape}$)$]. 
When all the \nl names are short and approximately equally long, we can roughly estimate the length (in words) of a sentence that will be produced to report a single fact, before actually producing the sentence, by counting the elements of the sentence plan that will be used to produce the sentence. Furthermore, we can roughly estimate the length of an aggregated sentence, i.e., a sentence that will be obtained by aggregating the simpler sentences (each reporting a single fact) of a group (bucket of Fig.~\ref{ILPModelChoices}), by counting the \emph{distinct} elements (no duplicates) of the sentence plans that will be used to produce the simple sentences of the group, because duplicate elements (originating from more than one simple sentences) are typically expressed only once in the aggregated sentence. 

In the following aggregation example, there are initially two simple sentences, produced by sentence plans identical to the first one of Section~\ref{sentencePlansBackground}, except for the different prepositions. The sentence plans of the two  simple sentences have four elements each: [$\mathit{ref}(S$ = \code{:BancroftChardonnay}$)$], [make]$_{\textit{present}, \; \textit{passive}}$, [by], [$\mathit{ref}(O$ = \code{:Mountadam}$)$] and [$\mathit{ref}(S$ = \code{:BancroftChardonnay}$)$], [make]$_{\textit{present}, \; \textit{passive}}$, [in], [$\mathit{ref}(O$ = \code{:Bancroft}$)$]. The distinct elements of the two sentence plans are only six, indicating that the aggregated sentence will be shorter than the two initial sentences together (eight elements in total). 

\begin{quoting}
{\small
\noindent
Bancroft Chardonnay is made by Mountadam. It is made in Bancroft.} 
$\Rightarrow$\\
{\small
Bancroft Chardonnay is made by Mountadam in Bancroft.}
\end{quoting}

\noindent By contrast, if a slightly different sentence plan involving the verb `produce' is used in the first simple sentence, the aggregated sentence will be longer, as shown below. The sentence plans of the two simple sentences again have eight elements in total, but their distinct elements are seven ([$\mathit{ref}(S$ = \code{:BancroftChardonnay}$)$], [produce]$_{\textit{present}, \; \textit{passive}}$, [by], [$\mathit{ref}(O$ = \code{:Mountadam}$)$], [make]$_{\textit{present}}$, [in],  [$\mathit{ref}(O$ = \code{:Bancroft}$)$]), correctly predicting that the aggregated sentence will now be longer. 

\begin{quoting}
{\small
\noindent
Bancroft Chardonnay is produced by Mountadam. It is made in Bancroft.} 
$\Rightarrow$\\
{\small
Bancroft Chardonnay is produced by Mountadam and made in Bancroft.}
\end{quoting}

\noindent The number of distinct elements is only an approximate estimate of the length of the aggregated sentence, because some of the names of the classes and individuals (e.g., `Bancroft Chardonnay') and some of the verb forms (e.g., `is made') are multi-word, but it allows the \ilp model to roughly predict the length of an aggregated sentence by considering only sentence plans, before actually producing or aggregating any sentences. 

The previous examples also show that selecting among alternative sentence plans affects the length of the generated text, not only because different sentence plans  may require more or fewer words to express the same fact, but also because different combinations of sentence plans may produce more or fewer aggregation opportunities (e.g., shared verbs). Content selection also affects the length of the text, not only because different facts may require more or fewer words to report, but also because the selected facts may or may not have combinations of sentence plans that provide aggregation opportunities, and the aggregation opportunities may allow saving fewer or more words. For example, consider the following facts. Let us assume that all four facts are equally important, and that we want to generate a text expressing only four of them.

{\small
\begin{verbatim}
   <:MountadamRiesling, isA, :Riesling>
   <:MountadamRiesling, :hasBody, :Medium>
   <:MountadamRiesling, :hasMaker, :Mountadam> 
   <:MountadamRiesling, :hasFlavor, :Delicate>
   <:MountadamRiesling, :hasSugar, :Dry>
\end{verbatim}
}

\noindent A pipeline approach to generation, where the content selection decisions are made greedily without considering their effects on the later stages of lexicalization (in our case, sentence plan selection) and aggregation, might select the first four of the facts (perhaps randomly, since all facts are equally important). Assuming that lexicalization also does not consider the effects of its choices (selected sentence plans) on sentence aggregation, we may end up with the following text, before and after aggregation.

\begin{quoting}
{\small 
\noindent This is a Riesling. It is medium. It is produced by Mountadam. It has a delicate flavor. 
$\Rightarrow$\\
This is a medium Riesling, produced by Mountadam. It has a delicate flavor.
}
\end{quoting}

\noindent On the other hand, a global approach that jointly considers the decisions of content selection, lexicalization, and aggregation might prefer to express the fifth fact instead of the fourth, and to use sentence plans that allow more compressive aggregations, leading to a much shorter text, as shown below.

\begin{quoting}
{\small
\noindent This is a Riesling. It is medium. It is dry. It is delicate.
$\Rightarrow$
This is a medium dry delicate Riesling.
}
\end{quoting}

The length of the resulting text is important when space is limited or expensive, as already discussed, which is why we aim to produce compact texts, i.e., texts that report as many facts per word as possible (or texts that maximize the importance of the reported facts divided by the words used, when facts are not equally important). More precisely, given an individual or class of an \owl ontology and a set of available facts about it, we aim to produce a text that:
\begin{description}
\item[Goal 1:] expresses as many of the available facts as possible (or a text that maximizes the total importance of the reported facts, when facts are not equally important), 
\item[Goal 2:] using as few words as possible.
\end{description}
By varying weights associated with Goals 1 and 2, we obtain different compact texts, aimed towards expressing more of the available facts at the expense of possibly using more words, or aimed towards using fewer words at the expense of possibly expressing fewer of the available facts. 

We can now formally define our first \ilp model. Let $s_1, \dots, s_m$ be disjoint subsets (buckets of Fig.~\ref{ILPModelChoices}) of $F = \{f_1, \dots, f_n\}$ (the set of available facts), each containing 0 to $n$ facts.
A single aggregated sentence is generated from each subset $s_{j}$ by aggregating the simple sentences (more precisely, their selected sentence plans) that express the facts of $s_j$. An empty $s_j$ generates no sentence. Hence, the resulting text can be at most $m$ aggregated sentences long. Let us also define:
\begin{eqnarray}
\label{aVariable}
a_{i} &=& \left\{
  \begin{array}{l l}
    1, & \text{if fact $f_{i}$ is selected}\\
    0, & \text{otherwise}\\
  \end{array} \right.
\\
\label{lVariable}
l_{ikj} &=& \left\{
  \begin{array}{l l}
    1, & \text{if sentence plan $p_{ik}$ is used to express fact $f_{i}$, 
           and $f_{i}$ is in subset $s_{j}$}\\
    0, & \text{otherwise}\\
  \end{array} \right.
\\
\label{bVariable}
b_{tj} &=& \left\{
  \begin{array}{l l}
    1, & \text{if element $e_{t}$ is used in subset $s_{j}$}\\
    0, & \text{otherwise}\\
  \end{array} \right.
\end{eqnarray}

\noindent and let $B$ be the set of all the distinct elements (no duplicates) from all the available sentence plans $p_{ik}$ that can express the facts of $F$. As already noted, the length of an aggregated sentence resulting from a subset $s_j$ can be roughly estimated by counting the distinct elements of the sentence plans that were chosen to express the facts of $s_j$. 

The objective function of our first \ilp model (Eq.~\ref{AUEB_NLG_ILP_CS_AGG} below) maximizes the total importance of the selected facts (or simply the number of selected facts, if all facts are equally important), and minimizes the number of distinct elements in each subset $s_{j}$, i.e., the approximate length of the corresponding aggregated sentence; an alternative explanation is that by minimizing the number of distinct elements in each $s_j$, we favor subsets that aggregate well. By $a$ and $b$ we jointly denote all the $a_{i}$ and $b_{tj}$ variables. $|\sigma|$ denotes the cardinality of a set $\sigma$. The two parts of the objective function are normalized to $[0, 1]$ by dividing by the total number of available facts $|F|$ and the number of subsets $m$ times the total number of distinct elements $|B|$. We  multiply $\alpha_i$ with the importance score $\mathit{imp}(f_i)$ of the corresponding fact $f_i$. We assume that the importance scores range in $[0,1]$; in our experiments, all the importance scores are set to $1$, with the exception of redundant message triples that are assigned zero importance scores (Section~\ref{InterestReasoning}). The parameters $\lambda_1, \lambda_2$ are used to tune the priority given to expressing many important facts vs.\ generating shorter texts; we set $\lambda_1$ + $\lambda_2$ = 1.

Constraint \ref{aConstraint} ensures that for each selected fact, exactly one sentence plan is selected and that the fact is placed in exactly one subset; if a fact is not selected, no sentence plan for the fact is selected and the fact is placed in no subset. In Constraint \ref{bConstraint}, $B_{ik}$ is the set of distinct elements $e_t$ of the sentence plan $p_{ik}$. This constraint ensures that if $p_{ik}$ is selected in a subset $s_{j}$, then all the elements of $p_{ik}$ are also present in $s_{j}$. If $p_{ik}$ is not selected in $s_{j}$, then some of its elements may still be present in $s_j$, if they appear in another selected sentence plan of $s_j$. In Constraint \ref{bConstraint2}, $P(e_t)$ is the set of sentence plans that contain element $e_{t}$. If $e_{t}$ is used in a subset $s_{j}$, then at least one of the sentence plans of $P(e_t)$ must also be selected in $s_{j}$. If $e_{t}$ is not used in $s_{j}$, then no sentence plan of $P(e_t)$ may be selected in $s_j$. Constraint \ref{lengthConstraint} limits the number of elements that a subset $s_{j}$ can contain to a maximum allowed number $B_{max}$, in effect limiting the maximum (estimated) length of an aggregated sentence. Constraint \ref{sectionConstraint} ensures that facts from different topical sections will not be placed in the same subset $s_j$, to avoid unnatural aggregations. 

\begin{equation}
\max_{a,b}{\lambda_1 \cdot \sum_{i=1}^{|F|}{\frac{a_{i} \cdot \mathit{imp}(f_{i})}{|F|}} 
- \lambda_2 \cdot \sum_{j=1}^{m}\sum_{t=1}^{|B|}{\frac{b_{tj}}{m \cdot |B|}}} 
\label{AUEB_NLG_ILP_CS_AGG}
\end{equation}
\noindent subject to:
\vspace{-7mm}
\begin{gather}
\label{aConstraint}
a_{i} = \sum_{j=1}^{m}\sum_{k=1}^{|P_{i}|}{l_{ikj}}, \mbox{for} \; i=1,\dots,n
\\
\label{bConstraint}
\sum_{e_{t} \in B_{ik}}{b_{tj}} \geq |B_{ik}| \cdot l_{ikj}, \mbox{for} \left\{
	\begin{array}{l}
    i=1,\dots,n \\
    j=1,\dots,m\\
    k=1,\dots,|P_{i}|\\
\end{array} \right.
\\
\label{bConstraint2}
\sum_{p_{ik} \in P(e_t)}{l_{ikj}} \geq b_{tj} , \mbox{for} \left\{
\begin{array}{l}
    t=1,\dots,|B|\\
    j=1,\dots,m\\
\end{array} \right.
\\
\label{lengthConstraint}
\sum_{t=1}^{|B|}{b_{tj}} \leq B_{max} , \mbox{for} \; j=1,\dots,m
\\
\label{sectionConstraint}
\sum_{k=1}^{|P_{i}|}{l_{ikj}} + \sum_{k'=1}^{|P_{i'}|}{l_{i'k'j}} \leq 1 , \mbox{for} \left\{
\begin{array}{l}
j=1,\dots,m, \; i = 2, \dots, n\\
i' = 1, \dots, n-1 ; i \neq i' \\
\textit{section}(f_i) \neq \textit{section}(f_i') \\
\end{array} \right.
\end{gather}

\subsection{Our Extended ILP Model} \label{OurExtendedModel}

The \ilp model of the previous section assumes that a single \nl name is available for each individual or class (excluding anonymous ones). By contrast, our extended \ilp model assumes that multiple alternative \nl names are available. The reader is reminded that an \nl name specifies how to generate a noun phrase naming an individual or class, and that it is a sequence of slots, along with instructions specifying how to fill them in. 

For an individual or class acting as the $O$ of a fact $\left<S, R, O\right>$ to be expressed, the extended \ilp model always selects the shortest available \nl name. It takes, however, into account the length of the (shortest) \nl name of $O$ when estimating the length of a sentence that will express $\left<S, R, O\right>$. By contrast, the model of the previous section ignored the lengths of the \nl names when estimating sentence lengths, assuming that all the \nl names are short and approximately equally long, an assumption that does not always hold. For example, the Disease Ontology, one of the ontologies of our experiments, includes an individual with an \nl name that produces the noun phrase ``paralysis of the legs due to thrombosis of spinal arteries'', and another individual with an \nl name that produces simply ``inflammation''. Hence, a sentence that uses the former \nl name to express a fact whose $O$ is the former individual will be much longer than a sentence that uses the latter \nl name to express another fact whose $O$ is the latter individual, even if both sentences are produced by the same sentence plan. 

The extended model also considers the possibility of $O$ being a conjunction or disjunction of classes, individuals,  datatype values (Section~\ref{NaturalOWL}), as in the last fact below.

{\small
\begin{verbatim}
 <:BrazilianHemorrhagicFever, :isA, :ViralInfectiousDisease>
 <:BrazilianHemorrhagicFever, :hasMaterialBasisIn, :SabiaVirus>
 <:BrazilianHemorrhagicFever, :transmittedBy, :rodents>
 <:BrazilianHemorrhagicFever, :hasSymptom,
							                              and(:fatigue, :muscleAches, :dizziness)>
\end{verbatim}
}

\noindent In the \ilp model of the previous section, we made no distinction between $O$s that are single classes, individuals, or datatype values, and $O$s that are conjunctions or disjunctions, assuming that the number of conjuncts or disjuncts, respectively, is always small and does not affect much the length of the resulting sentence. In some ontologies, though, the number of conjuncts or disjuncts varies greatly. In the Disease Ontology, the number of conjuncts in 
the \code{hasSymptom} relation ranges from 1 to 14. Let us assume that we wish to generate a text for 
\code{BrazilianHemorrhagicFever}, that we are limited to expressing two facts, and that all facts are equally important. The model of the previous section might, for example, select the first and last of the facts above, possibly because their sentence plans are short (in elements), leading to the following sentence. 

\begin{quoting}
{\small
\noindent The Brazilian hemorrhagic fever is a viral infectious disease that causes fatigue, muscle aches and dizziness.}
\end{quoting}

\noindent By contrast, the extended \ilp model takes into account that the conjunction in the $O$ of the last fact above requires 
five words. Hence, it might select the first and 
third facts instead, producing the following shorter sentence. 

\begin{quoting}
{\small
\noindent 
The Brazilian hemorrhagic fever is a viral infectious disease transmitted by rodents.
}
\end{quoting}

\noindent Note, also, that selecting the first and 
second facts, which only have single individuals or classes as $O$s, would lead to the following sentence, which is longer, because of the length of 
``the Sabia virus''.  

\begin{quoting}
{\small
\noindent 
The Brazilian hemorrhagic fever is a viral infectious disease caused by the Sabia virus.}
\end{quoting}

Selecting among the alternative \nl names of the $S$ of a fact $\left<S,R,O\right>$ is more complicated, because a longer \nl name (e.g., producing ``the Napa Region Bancroft Chardonay wine'') may also convey some of the other available facts, without requiring separate sentences for them, thus saving words. Consider, for example, the following facts and assume that we wish to generate a text expressing all of them.

{\small
\begin{verbatim}
   <:BancroftChardonnay, isA, :Chardonnay>
   <:BancroftChardonnay, :locatedIn, :NapaRegion>
   <:BancroftChardonnay, :hasMaker, :Bancroft>
   <:BancroftChardonnay, :hasFlavor, :Moderate>
   <:BancroftChardonnay, :hasSugar, :Dry>
\end{verbatim}
}

\noindent Let us also assume that \code{BancroftChardonnay} has three alternative \nl names, which produce ``Bancroft Chardonnay'', ``the Napa Region Bancroft Chardonnay wine'', and ``the moderate tasting and dry Bancroft Chardonnay wine'', respectively.\footnote{Some of the \nl names of this article, like the first two of this example, were semi-automatically constructed by the methods of Lampouras \shortcite{Lampouras2015}. The other \nl names, like the third one of this example, were manually authored to provide more choices to the extended \ilp model.} For each alternative \nl name of $S$, we invoke the mechanism of \nlowl (Section~\ref{InterestReasoning}) that detects redundant facts (message triples with zero importance scores). In our example, if we choose to refer to $S$ as ``Bancroft Chardonnay'', we do not need to produce separate sentences for the first and third facts above, since they are already indirectly expressed by the \nl name of $S$, and similarly for the other two \nl names of $S$, as shown below. 

\begin{quoting}
{\small 
\noindent $S$ called ``Bancroft Chardonnay'':\\
Bancroft Chardonnay is moderate and dry. It is produced in the Napa Region. \\
\sout{It is a Chardonnay. It is produced by Bancroft.}}
\end{quoting}
\begin{quoting}
{\small
\noindent $S$ called ``the Napa Region Bancroft Chardonnay wine'':\\
The Napa Region Bancroft Chardonnay wine is moderate and dry.\\
\sout{It is a Chardonnay. It is produced by Bancroft in the Napa Region.}}
\end{quoting}
\begin{quoting}
{\small
\noindent $S$ called ``the moderate tasting and dry Bancroft Chardonnay wine'':\\
The moderate tasting and dry Bancroft Chardonnay wine is produced in the Napa Region.\\
\sout{It is a moderate, dry Chardonnay. It is produced by Bancroft.}}
\end{quoting}

\noindent Selecting the \nl name that produces the
shortest noun phrase (``Bancroft Chardonnay'') does not lead to the shortest text. The shortest text is obtained when the second \nl name is selected. Selecting the third \nl name above, which leads to the largest number of facts made redundant (meaning facts that no longer need to be expressed as separate sentences), also does not lead to the shortest text, as shown above. 

To further increase the range of options that the extended \ilp model considers and help it to produce more compact texts, when using the extended \ilp model we allow alternative \nl names to be provided also for individuals or classes declared as `anonymous' (Section~\ref{NLnamesBackground}); \emph{possibly anonymous} is now a better term. 
In other words, the system can refer to an individual or class declared to be possibly anonymous, by using a demonstrative pronoun (``this'') or a demonstrative noun phrase mentioning the parent class (e.g., ``this Chardonnay''), as with anonymous individuals and classes before, but it can also use an \nl name of the individual or class (if provided), i.e., declaring an individual or class as possibly anonymous licenses the use of a demonstrative or demonstrative noun phrase, without excluding the use of an \nl name.\footnote{A pronoun can also be used, but pronouns are generated by the (external to our \ilp models) referring expression generation component of \nlowl, after invoking the \ilp models, as already discussed.} Continuing our example, let us assume that \code{BancroftChardonnay} has been declared as possibly anonymous. Then the following texts are also possible. 

\begin{quoting}
{\small
\noindent Demonstrative used for $S$: \\
This is a moderate, dry Chardonnay. It is produced by Bancroft in the Napa Region.}
\end{quoting}
\begin{quoting}
{\small
\noindent Demonstrative noun phrase used for $S$: \\
This Chardonnay is moderate and dry. It is produced by Bancroft in the Napa Region.\\
\sout{It is a Chardonnay.}}
\end{quoting}

\noindent As illustrated above, a demonstrative noun phrase that mentions the ancestor class (e.g., ``this Chardonnay'') is also taken to express the corresponding fact about the ancestor class (e.g.,  \code{<:BancroftChardonnay, isA, :Chardonnay>}). Notice, also, that using a demonstrative or demonstrative noun phrase does not necessarily lead to the shortest text. In our example, the shortest text is still obtained using the second \nl name.

Before moving on to the formulation of the extended \ilp model, let us discuss how it estimates the lengths of (possibly aggregated) sentences. In the \ilp model of the previous section, we roughly estimated the length of an aggregated sentence resulting from a subset (bucket) $s_j$ by counting the distinct elements of the sentence plans chosen to express the facts of $s_j$. For example, let us assume that the distinct elements [$\mathit{ref}(S$ = \code{:StEmilion}$)$], [make]$_{\textit{present}, \; \textit{passive}}$, [from], and [$\mathit{ref}(O$ = \code{:cabernetSauvignonGrape}$)$] are used in a single subset $s_j$. The \ilp model of the previous section did not consider the lengths of the noun phrases that will be produced by the \nl names of \code{:StEmilion} and \code{:cabernetSauvignonGrape} of the elements [$\mathit{ref}(S$ = \code{:StEmilion}$)$] and [$\mathit{ref}(O$ = \code{:cabernetSauvignonGrape}$)$]. 
Also, it did not take into account that the element [make]$_{\textit{present}, \; \textit{passive}}$ actually produces two words (``is made''). 

The extended model defines a function $length(e_{t})$ that maps each distinct element $e_{t}$ to the length (in words) of the text it produces (e.g., ``is made''). More specifically, if $e_{t}$ is an element referring to a single individual or class acting as the $O$ of a message triple (e.g., [$\mathit{ref}(O$ = \code{:cabernetSauvignonGrape}$)$]), then $length(e_{t})$ is the length (in words) of the (shortest) \nl name of $O$; if $O$ is a conjunction or disjunction, then $length(e_{t})$ is the sum of the lengths of the (shortest) \nl names of all the conjuncts or disjuncts. However, if $e_{t}$ is an element referring to $S$ (e.g., [$\mathit{ref}(S$ = \code{:StEmilion}$)$]), then $length(e_{t}) = 1$, because the \nl name of $S$ will be used only once at the beginning of the text, and each subsequent reference to $S$ will be via a pronoun of length 1 (e.g., ``St. Emilion is red and strong. \emph{It} is made from Cabernet Sauvignon grapes.''); the first occurrence of the \nl name is counted separately, directly in the objective function discussed below. The estimated length of a (possibly aggregated) sentence is the sum of the estimated lengths ($\sum_{t} length(e_{t})$) of the distinct elements of the sentence plan(s) that produced it. Overall, the extended model estimates more accurately the length of the text that will be produced, though the actual text length may still be slightly different; for example, connectives or complementizers (e.g., `and', `that') may be added during aggregation.

We can now formally define our extended \ilp model. As in the simpler model of Section~\ref{OurModel}, $F$ is the set of available facts $f_i$ about the individual or class $S$ we wish to generate a text for, and $s_1, \dots, s_m$ are disjoint subsets of $F$ (buckets of Fig.~\ref{ILPModelChoices}) showing which simple sentences (each expressing a single fact of $F$) will be aggregated together. Let $N = \{n_{1}, n_{2}, \dots\}$ be a set of alternative \nl names for $S$. Recall that we model only the choice of \nl name for $S$, assuming that the shortest \nl name is always used for the $O_i$ of each fact $f_i = \left<S,R_i,O_i\right>$. Each $a_{i}$ variable now indicates if the corresponding fact $f_i$ is explicitly expressed by generating a sentence:
\begin{equation}\label{aVariable_extend}
a_{i} = \left\{
  \begin{array}{l l}
    1, & \text{if the fact $f_{i}$ is expressed as a sentence}\\
    0, & \text{otherwise}\\
  \end{array} \right.
\end{equation}

\noindent By contrast, $d_i$ is more general; $d_i = 1$ if the corresponding fact $f_i$ is conveyed either explicitly (by generating a sentence for $f_i$) or implicitly (via an \nl name): 
\begin{equation}\label{dVariable}
d_{i} = \left\{
  \begin{array}{l l}
    1, & \text{if the fact $f_{i}$ is expressed as a sentence or via an \nl name }\\
    0, & \text{otherwise}\\
  \end{array} \right.
\end{equation}

\noindent The distinction between $a_i$ and $d_i$ is necessary, because when  a fact $f_{i}$ is expressed as a sentence, a sentence plan for $f_i$ is also selected. For example, a fact $f_{i} =$ <\code{:BancroftChardonnay}, \code{:hasMaker}, \code{:Bancroft}> can be expressed as a sentence in the final text (e.g., ``This \emph{is produced by Bancroft}. It comes from the Napa Region.'') or through an \nl name (e.g., ``\emph{Bancroft Chardonnay} is produced in the Napa Region.''). In both texts, $f_{i}$ is  expressed ($d_i = 1$), but in the former text $a_i = 1$, whereas in the latter one $a_i = 0$.

The $l_{ikj}$ and $b_{tj}$ variables are as in the \ilp model of the previous section (Eq.~\ref{lVariable} and \ref{bVariable}). 
For the extended model, we also define:
\begin{equation}\label{mVariable}
m_{r} = \left\{
  \begin{array}{l l}
    1, & \text{if the \nl name $n_{r}$ is used for $S$}\\
    0, & \text{otherwise}\\
  \end{array} \right.
\end{equation}

Similarly to the previous model's objective function (\ref{AUEB_NLG_ILP_CS_AGG}), the extended model's objective function (\ref{AUEB_NLG_ILP_CS_AGG_EXTEND}) maximizes the total importance of the expressed facts (or simply the number of expressed facts, if all facts are equally important), and minimizes the length of the distinct elements in each subset $s_{j}$ and the length of the (single, initial occurrence of the) \nl name used to express $S$, i.e., the approximate length of the resulting text. By $d$, $b$, and $m$ we jointly denote all the $d_{i}$, $b_{tj}$, and $m_{r}$ variables. The left part of the objective is the same as in the previous model, with the variables $a_{i}$ replaced by $d_i$. In the right part, we multiply the $b_{tj}$ and $m_{r}$ variables with the functions $length(e_{t})$ and $length(n_{r})$, which calculate the lengths (in words) of the corresponding element ($e_t$) and \nl name ($n_r$), respectively. The two parts of the objective function are normalized to $[0, 1]$ by dividing by the total number of available facts $|F|$ and the number of subsets $m$ times the total length of distinct elements $|B|$ plus the total length of the $R$ available \nl names. Again, the parameters $\lambda_1, \lambda_2$ are used to tune the priority given to expressing many important facts vs.\ generating shorter texts; we set $\lambda_1$ + $\lambda_2$ = 1.
\begin{equation}
\max_{d,b,m}{\lambda_1 \cdot \sum_{i=1}^{|F|}{\frac{d_{i} \cdot \mathit{imp}(f_{i})}{|F|}} 
- \lambda_2 \cdot 
(\frac{\sum_{j=1}^{m}\sum_{t=1}^{|B|}{b_{tj} \cdot length(e_{t})} + \sum_{r=1}^{|R|}{m_{r} \cdot length(n_{r})}}
{m \cdot \sum_{t=1}^{|B|}{length(e_{t})} + \sum_{r=1}^{|R|}{length(n_{r})}})} 
\label{AUEB_NLG_ILP_CS_AGG_EXTEND}
\end{equation}
\noindent subject to:
\vspace*{-7mm}
\begin{gather}
\label{aConstraintExtend}
a_{i} = \sum_{j=1}^{m}\sum_{k=1}^{|P_{i}|}{l_{ikj}}, \mbox{for} \; i=1,\dots,n
\\
\label{bConstraintExtend}
\sum_{e_{t} \in B_{ik}}{b_{tj}} \geq |B_{ik}| \cdot l_{ikj}, \mbox{for} \left\{
\begin{array}{l}
    i=1,\dots,n \\
    j=1,\dots,m\\
    k=1,\dots,|P_{i}|\\
\end{array} \right.
\\
\label{bConstraint2Extend}
\sum_{p_{ik} \in P(e_t)}{l_{ikj}} \geq b_{tj} , \mbox{for} \left\{
\begin{array}{l}
    t=1,\dots,|B|\\
    j=1,\dots,m\\
\end{array} \right.
\\
\label{lengthConstraintExtend}
\sum_{t=1}^{|B|}{b_{tj} \cdot length(e_{t})} \leq W_{max} , \mbox{for} \; j=1,\dots,m
\\
\label{sectionConstraintExtend}
\sum_{k=1}^{|P_{i}|}{l_{ikj}} + \sum_{k'=1}^{|P_{i'}|}{l_{i'k'j}} \leq 1 , \mbox{for} \left\{
\begin{array}{l}
j=1,\dots,m, \; i = 2, \dots, n\\
i' = 1, \dots, n-1 ; i \neq i' \\
\textit{section}(f_i) \neq \textit{section}(f_i') \\
\end{array} \right.
\end{gather}
\begin{gather}
\label{mConstraintExtend}
\sum_{r=1}^{|N|}{m_{r}} = 1
\\
\label{admConstraintExtend}
d_{i} = a_i + \sum_{m_{r} \in R(f_{i})}{m_{r}}, \mbox{for} \; i=1,\dots,n
\end{gather}

Constraints \ref{aConstraintExtend}--\ref{sectionConstraintExtend} serve the same purpose as in the previous model (Eq.~\ref{aConstraint}--\ref{sectionConstraint}), except that Constraint~\ref{lengthConstraintExtend} now limits the number of words (instead of elements) that a subset $s_{j}$ can contain to a maximum allowed number $W_{max}$. Constraint \ref{mConstraintExtend} ensures that exactly one \nl name is selected from the available \nl names of $S$. In Constraint \ref{admConstraintExtend}, $R(f_{i})$ is the set of \nl names that (indirectly) express the 
fact $f_{i}$. If $f_i$ is to be expressed (i.e., $d_{i}=1$), then either one of the \nl names in $R(f_{i})$ must be selected, or a sentence for $f_i$ must be generated ($a_i = 1$), not both. If $f_i$ is not to be expressed, then none of the \nl names in $R(f_{i})$ may be selected, nor should a sentence be generated for $f_{i}$.

\section{Computational Complexity and Approximations} \label{Complexity}

The models of Sections \ref{OurModel} and \ref{OurExtendedModel} are formulated as \ilp problems, more precisely binary \ilp problems since all their variables are binary. Solving binary \ilp problems is in general 
\textsc{np}-hard \cite{Karp72}. 
We also note that content selection, as performed by our models, is similar to the 0-1 multiple Knapsack problem, which 
is also \textsc{np}-hard. In both 
cases, we have 
$n$ items (facts),
$m$ knapsacks (fact 
subsets, buckets) of a certain capacity, and we wish to 
fill the knapsacks with $m$ disjoint 
subsets of 
the available items, so that the total importance of the selected 
items (items placed in the knapsacks) is maximum. However, in our models each 
item (fact) is further associated with a set of (sentence plan) elements, subsets of which are possibly 
shared (in a subset, bucket) with other 
items (facts),
and the capacity of the knapsacks is specified in distinct elements. Furthermore, the elements of each item depend on the selected sentence plans,  
there are additional constraints to comply with topical sections, and the objective function of our models also tries to minimize the total length of the resulting text.
Hence, our models do not correspond directly to the 0-1 multiple Knapsack problem.

A possible approach to solve \ilp models in polynomial time is to relax the constraint that 
variables are integer (or binary) and solve the resulting Linear Programming model (\textsc{lp} relaxation) using, for example, the Simplex algorithm \cite{Dantzig1963}. The resulting values of the variables are then rounded to the closest integral values. The 
solution is not guaranteed to be optimal for the original \ilp problem, nor feasible (some constraints of the original problem may be violated). The solution of the \textsc{lp} relaxation, though, is 
the same as the solution of the original \ilp problem if the problem can be formulated as $\max_x c^{T}x$ with constraints $Ax = b$, where $c$, $A$, $m$ have integer values and the matrix $A$ is totally unimodular \cite{Schrijver86,Roth2004}. 
An integer matrix is totally unimodular if every square, nonsingular submatrix is unimodular (i.e., its determinant is 0, 1, or -1). Unfortunately, this is not the case in our \ilp models.

In practice, 
off-the-shelf solvers that solve the original \ilp problem (not the \textsc{lp} relaxation) are very fast when the number of variables is small.\footnote{We use the branch-and-cut implementation of \textsc{glpk} with mixed integer rounding, mixed cover, and clique cuts; see \url{http://sourceforge.net/projects/winglpk/}.} Our experiments show that solving the first \ilp model is reasonably fast, provided that the number of fact subsets (buckets) is $m \leq 4$. Indeed, $m$ seems to be the greatest factor to the model's complexity; the number of variables in the model grows exponentially to $m$, while the effect of the other parameters (e.g., number of available facts $|F|$) is weaker. We did not examine experimentally how the solving times of the extended \ilp model relate to the number of subsets $m$; however, 
the variables in the extended model 
also grow exponentially to the number of fact subsets $m$.

\begin{figure}
\center
\includegraphics[width=\columnwidth]{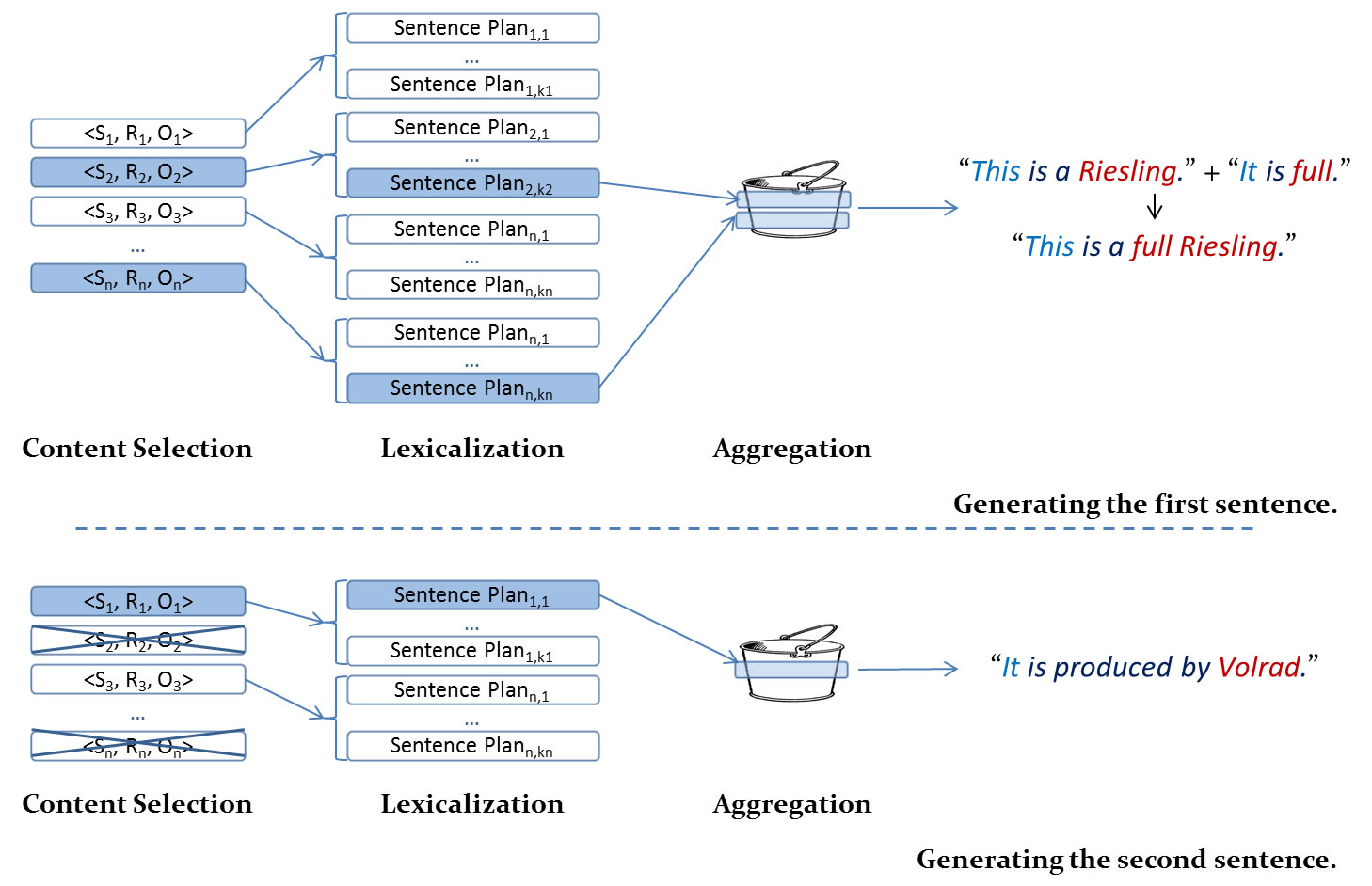}
\caption{Illustration of the approximation of the first \ilp model.}
\label{ILPModelApprox}
\end{figure}

When the number of variables is too large to solve the first \ilp model efficiently, we use an approximation of the model, which considers each fact subset (bucket, aggregated sentence of the final text) separately
(Fig.~\ref{ILPModelApprox}). We start with the full set of available facts ($F$) and use the first \ilp model with $m=1$ to produce the first (aggregated) sentence of the final text. We then remove the facts expressed by the first (aggregated) sentence from $F$, and use the \ilp model, again with $m=1$, to produce the second (aggregated) sentence etc. This process is repeated until we produce the maximum number of allowed aggregated sentences, or until we run out of available facts.  

Since the approximation of the first \ilp model does not consider all the fact subsets jointly, it does not guarantee finding a globally optimal solution for the entire text. Nevertheless, experiments (presented below) that compare the approximation to the original first \ilp model show no apparent decline in text quality nor in the ability to produce compact texts. 
Solving times now grow almost linearly to both the number of subsets $m$ and the number of available facts $|F|$. Furthermore, $|F|$ decreases in every subsequent solving of the model (to produce the next aggregated sentence of the text), which reduces the time needed by the solver. Our experiments indicate that the approximation can guarantee practical running times even for $m \geq 5$, while still outperforming the pipeline approach in terms of producing more compact texts.  

The same approximation (considering each fact subset separately) can be applied to our extended \ilp model. We did not experiment with the approximation of the extended model, however, because the only ontology we considered that required $m \geq 5$ and, hence, an approximation (Consumer Electronics Ontology) did not require the extended model (the lengths of the \nl names did not vary significantly, and we could not think of alternative \nl names for the products being described).

\section{Experiments} \label{Experiments}

We now present the experiments we performed to evaluate our \ilp models. We first discuss the ontologies and systems that were used in our experiments.

\subsection{The Ontologies of our Experiments} \label{ontologiesOfExperiments}

We experimented with three \owl ontologies: (1) the Wine Ontology, which provides information about wines, wine producers etc.; (2) the Consumer Electronics Ontology, intended to help exchange information about consumer electronics products; and (3) the Disease Ontology, which describes diseases, including their symptoms, causes etc.\footnote{Consult  \url{http://www.w3.org/TR/owl-guide/wine.rdf/}, \url{http://www.ebusinessunibw.org/ontologies/consumerelectronics/v1}, and \url{http://disease-ontology.org/}.}  
The Wine Ontology is one of the most commonly used examples of \owl ontologies and involves a wide variety of \owl constructs; hence, it is a good test case for systems that produce texts from \owl. The Consumer Electronics and Disease Ontologies were constructed by biomedical and e-commerce experts to address real-life information needs; hence, they constitute good real-world test cases from different domains. 

The Wine Ontology contains 63 wine classes, 52 wine individuals, a total of 238 classes and individuals (including wineries, regions, etc.), and 14 relations (properties). Manually authored, high-quality domain-dependent generation resources (text plans, sentence plans, \nl names etc.) for \nlowl are available for this ontology from our previous work \cite{Androutsopoulos2013}. 

The Consumer Electronics Ontology 
comprises 54 classes and 441 individuals (e.g., printer types, paper sizes, manufacturers), but no information about particular products. In previous work \cite{Androutsopoulos2013}, we added 60 individuals (20 digital cameras, 20 camcorders, 20 printers). The 60 individuals were randomly selected from a publicly available dataset of 286 digital cameras, 613 camcorders, and 58 printers that complies with the Consumer Electronics Ontology.\footnote{The dataset was obtained from 
\url{http://rdf4ecommerce.esolda.com/}. A list of similar datasets is
available at \url{http://wiki.goodrelations-vocabulary.org/Datasets}.}
From these 60 individuals, we generate texts for the 30 `development' individuals (10 cameras, 10 camcorders, 10 printers), for which high-quality manually authored domain-dependent generation resources are available from our previous work. 

The Disease Ontology currently contains information about 6,286 diseases, all represented as classes. Apart from \textsc{is-a} relations, synonyms, and pointers to related terms, however, all the other information is represented using strings containing quasi-English sentences with relation names used mostly as verbs. For example, there is an axiom in the ontology stating that the Rift Valley Fever (\textsc{doid}\_1328) is a kind of viral infectious disease (\textsc{doid}\_934). All the other information about the Rift Valley Fever is provided in a string, shown below as `Definition'. The tokens that contain underscores (e.g., \code{results\_in}) are relation names. The ontology declares all the relation names, but uses them only inside `Definition' strings. Apart from diseases, it does not define any of the other entities mentioned in the `Definition' strings (e.g., symptoms, viruses).

\begin{quoting}
{\small \noindent Name: Rift Valley Fever (\textsc{doid}\_1328)\\
\textsc{is-a}: viral infectious disease (\textsc{doid}\_934) \\
Definition: A viral infectious disease that \code{results\_in} infection, \code{has\_material\_basis\_in} Rift Valley fever virus, which is \code{transmitted\_by} Aedes mosquitoes. The virus affects domestic animals (cattle, buffalo, sheep, goats, and camels) and humans. The infection \code{has\_symptom} jaundice, \code{has\_symptom} vomiting blood, \code{has\_symptom} passing blood in the feces, \code{has\_symptom} ecchymoses (caused by bleeding in the skin), \code{has\_symptom} bleeding from the nose or gums, \code{has\_symptom} menorrhagia and \code{has\_symptom} bleeding from venepuncture sites. 
}
\end{quoting}

We defined as individuals all the non-disease entities mentioned in the `Definition' strings, also adding statements to formally express the relations mentioned in the original `Definition' strings. For example, the resulting ontology contains the following definition of Rift Valley Fever, where \code{:infection}, \code{:Rift\_Valley\_fever\_virus}, \code{:Aedes\_mosquitoes}, \code{:jaundice} etc.\ are new individuals. 

\label{DOID1328}
{\footnotesize
\begin{verbatim}
   SubClassOf(:DOID_1328 
      ObjectIntersectionOf(:DOID_934
         ObjectHasValue(:results_in :infection)
         ObjectHasValue(:has_material_basis_in :Rift_Valley_fever_virus)
         ObjectHasValue(:transmitted_by :Aedes_mosquitoes)
         ObjectHasValue(:has_symptom :jaundice)
         ObjectHasValue(:has_symptom :vomiting_blood)
         ObjectHasValue(:has_symptom :passing_blood_in_the_feces)
         ObjectHasValue(:has_symptom 
                        :ecchymoses_(caused_by_bleeding_in_the_skin))
         ObjectHasValue(:has_symptom :bleeding_from_the_nose_or_gums)
         ObjectHasValue(:has_symptom :menorrhagia)
         ObjectHasValue(:has_symptom :bleeding_from_venepuncture_sites)))
\end{verbatim}
}

\noindent The new form of the ontology was produced automatically, using patterns that searched the definition strings for relation names (e.g., \code{results\_in}), sentence breaks, and words introducing secondary clauses (e.g., ``that'', ``which'').\footnote{The 
new form of the Disease Ontology that we produced is available upon request and will be made publicly available when this article is published.} Some sentences of the original definition strings that did not include declared relation names (e.g., ``The virus affects\dots and humans'' in the `Definition' string of Rift Valley Fever) were discarded, because they could not be automatically converted to appropriate \owl statements.

The new form of the Disease Ontology contains 6,746 classes, 15 relations, and 1,545 individuals. From the 6,746 classes
(all describing diseases), 5,014 classes participate only in \textsc{is-a} and synonym relations; hence, texts for them would not be particularly interesting. From the remaining 
1,732 classes, we generate texts for the 200 randomly selected `development' classes of Evaggelakaki \shortcite{Evaggelakaki2014}, for which manually authored domain-dependent generation resources for \nlowl are available.

\subsection{The Systems of our Experiments} \label{experimentOverview} \label{systemsOverview}

We call \pipeline the original \nlowl, which uses a pipeline architecture. Two modified versions of \nlowl, called \ilpnlg and \ilpnlgextend, use our first and extended \ilp models, respectively. All the systems of our experiments share the same linguistic resources (e.g., text plans, sentence plans, \nl names, aggregation rules), ontologies, and importance scores; all facts are assigned an importance of $1$, except for facts that are automatically assigned zero importance scores (Section~\ref{InterestReasoning}). 

\pipeline has a parameter $M$ specifying the number of facts to report per generated text. During content selection, \pipeline ranks all the available facts ($F$) by decreasing importance, and selects the $M$ most important ones (or all of them if $M > |F|$) selecting randomly among facts with the same importance when needed.
In the experiments that follow, we generated texts with \pipeline for different values of $M$. For each $M$ value, the texts of \pipeline were generated 
$T$ times, each time using a different (randomly selected) alternative sentence plan of each relation,
and a different (randomly selected) \nl name of each individual or class (when multiple alternative \nl names were available). 
For the \pipeline model, we assume that the sentence plans and \nl names are uniformly distributed with each being equally probable to be selected.
For the aggregation of the selected facts, \pipeline uses the text planner from the original \nlowl. 
The text planner is invoked after content selection to partition the selected facts into topical sections, and to order the topical sections and the facts within each topical section. The aggregation rules are then applied to all the facts of each topical section (also considering their selected sentence plans). From the $T$ generated texts, \pipeline returns the one which is estimated to have the highest facts per word ratio. Rather than use the actual length of each produced text to calculate the facts per words ratio, the number of words is instead estimated as the sum of distinct elements in each sentence of the text, to better align the objective of \pipeline to that of \ilpnlg.

We also generated the texts (for different values of $M$) using a variant of \pipeline, 
dubbed \pipelinestoch, which selects randomly amongst available facts, in addition to sentence plans and \nl names. However, unlike \pipeline, the probability of each sentence plan or \nl name is based on their respective length (in distinct elements), with the shortest ones being more probable to be selected. The fact's probabilities are similarly estimated by the length of the shortest sentence plan and \nl name available to them. In regards to aggregation, \pipelinestoch constructs fact subsets (corresponding to sentences in the final text) with the objective of minimizing the number of distinct elements in each subset, similarly to \ilpnlg. Each subset is initialized with random facts (sampled based on the length of their available resources) and subsequent facts are randomly placed in each subset, with probabilities estimated on the number of elements each fact has in common with the facts already in that particular subset. As with \pipeline, for each $M$ the texts are generated $T$ times, and the one with the highest facts per word ratio is used for the evaluation.

A greedier variant of \pipeline, 
\pipelineshort
always selects the shortest (in elements) sentence plan among the available ones
and the shortest (in words) \nl name. 
In \pipelineshort, if a subset of facts has the same importance, they are additionally ranked by increasing length of the shortest sentence plan and \nl name available to each; this way the fact with the potential to generate the shortest sentence will be selected first. 

Our final baseline, \pipelinebeam extends the output of \pipelineshort by employing beam search to select alternative facts, sentence plans, \nl names and fact subsets. During content selection, \pipelinebeam selects the subset of $M$ facts with the shortest sentence plans and \nl names available to them (similarly to \pipelineshort), and subsequently replaces a single random (based on the length of the available resources) fact from this subset with a random non-selected fact. This process is repeated until $K-1$ additional fact subsets are constructed; all differing from the initial subset by one (replaced) fact. In a similar way, $K$ different sentence plan assignments, $K$ different \nl name assignment and $K$ different fact subset assignments are also constructed, differing from the respective assignments of \pipelineshort by one substitution each. The combination of these assignments result in $K$ $\times$ $K$ $\times$ $K$ $\times$ $K$ different texts for each $M$. As in the other baselines, the text amongst these with the highest estimated facts per words ratio is used for the evaluation. 

To better compare the output of the pipeline baselines, we set the number of generated texts $T$ that \pipeline, \pipelinestoch and \pipelineshort generate to $K$ $\times$ $K$ $\times$ $K$ $\times$ $K$ as \pipelinebeam.

All the systems use the same text planner (from the original \nlowl)
which is invoked 
before content selection to partition the 
facts into topical sections, and to order the topical sections and the facts within each topical section. 
Each of the systems described above have different strategies to partition the selected facts after content selection in sentences. The selected facts retain the order given from the text planner, and the sentences inherit the minimum order of their included facts. Afterwards, aggregation rules are applied to all the facts of each fact subset (also considering their selected sentence plans). 
the text planner is first invoked (before using the \ilp models) to partition all the available facts ($F$) into topical sections. It is also invoked after using one of the \ilp models, to order the sentences in each group (bucket) that the \ilp model has decided to aggregate; as already noted, the aggregation rules presuppose that the sentences to be aggregated are already ordered, which is why the text planer is invoked at this point. After applying the aggregation rules to each group of (ordered) sentences, \ilpnlg and \ilpnlgextend invoke the text planner again to order the topical sections and the (now aggregated) sentences within each topical section.

\ilpnlg assumes that there is a single \nl name per individual or class (excluding anonymous ones) and, hence, cannot be used when multiple alternative \nl names are available. By contrast, \ilpnlgextend can handle multiple alternative \nl names. For each text, it selects a single \nl name per individual and class (as discussed in Section~\ref{OurExtendedModel}), which is then replaced by a demonstrative, demonstrative noun phrase, or pronoun, whenever the referring expression generation component of the original \nlowl decides to. 
\pipeline and \pipelineshort can also handle multiple \nl names, but \pipeline selects randomly among the alternative \nl names, and \pipelineshort selects always the shortest one. Like \ilpnlgextend, for each text \pipeline and \pipelineshort select a single \nl name per individual and class, which is then replaced by a demonstrative, demonstrative noun phrase, or pronoun, whenever the referring expression generation component of the original \nlowl decides to. 

A variant of \pipelineshort, called \pipelineshortnln, always selects the shortest (now in words) sentence plan among the available ones, and the \nl name of $S$ (the individual or class the text is generated for) that indirectly expresses the largest number of available facts $f_i = \left<S,R_i,O_i\right>$ (Section~\ref{InterestReasoning}), thus not requiring sentences to express them.\footnote{Selecting the \nl name of $S$ that expresses the largest number of available facts usually leads to better facts per word ratios than simply selecting the shortest (in words) \nl name of $S$. If several \nl names of $S$ express the same number of available facts, \pipelineshortnln selects the shortest (in words) \nl name.}
For $O_i$, \pipelineshortnln selects the same (shortest in words) \nl name as \ilpnlgextend and \pipelineshort. Otherwise,  \pipelineshortnln is identical to \pipelineshort. \pipelineshortnln is a more appropriate baseline for \ilpnlgextend than \pipelineshort, because like \ilpnlgextend it estimates the lengths of sentences and \nl names in words, and it takes into account that \nl names may indirectly express some of the available facts. 

Finally, \ilpnlgapprox denotes a system that is identical to \ilpnlg (it uses our first \ilp model), but with the approximation of Section~\ref{Complexity}, whereby each (possibly aggregated) sentence of the text is generated separately.

\subsection{Overview of the Experiments}

Before presenting the details of our experiments, let us first provide an overview. We started by comparing \ilpnlg to \pipeline and \pipelineshort on the Wine Ontology, where experiments showed that \ilpnlg leads to more compact texts, i.e., texts with higher facts per word ratios, with no deterioration in the perceived quality of the resulting texts, compared to the texts of \pipeline and \pipelineshort. 

We then tried to repeat the same experiments on the Consumer Electronics Ontology, but \ilpnlg was too slow in many cases, because of the larger number of available facts per product ($|F|$) and the larger ($m=10$) number of subsets (buckets) required to express all (or many) of the available facts. 
To address this problem, we developed the approximation (Section~\ref{Complexity}) of \ilpnlg, which is used in \ilpnlgapprox. The approximation was much more efficient and achieved higher facts per word ratios than \pipeline and \pipelineshort, with no deterioration in the perceived quality of the texts. In texts expressing many facts, the perceived quality of the texts of \ilpnlgapprox was actually higher, comparing to the texts of \pipeline and \pipelineshort. 

We then moved on to the Disease Ontology, to experiment with an additional domain. Since the Disease Ontology only required $m=4$ fact subsets to express all the available facts per disease, \ilpnlgapprox was not required, and \ilpnlg was used instead. We found that \ilpnlg did not always perform better than \pipeline and \pipelineshort (in terms of facts per word ratios), because the lengths of the \nl names of the Disease Ontology vary a lot, and there are also several facts $\left<S,R,O\right>$ whose $O$ is a conjunction, sometimes with many conjuncts. To address these issues, we extended \ilpnlg to \ilpnlgextend, which consistently produced more compact texts than \pipeline and \pipelineshortnln on the Disease Ontology.

Lastly, we returned to the Wine Ontology to see how \ilpnlgextend performs with multiple alternative \nl names. For this experiment, we created alternative \nl names for the individuals and classes of the Wine Ontology; we could not do the same for the Consumer Electronics and Disease Ontologies, because the names of electronic products tend to be unique and we did not have the expertise to create alternative names of diseases. Indeed, \ilpnlgextend produced more compact texts than \pipeline and \pipelineshortnln from the Wine Ontology, when multiple \nl names were available.

\subsection{Experiments with the Wine Ontology} \label{wineExperiments}

In a first set of experiments, we used the Wine Ontology, along with the manually authored domain-dependent generation resources (e.g., text plans, \nl names, sentence plans) we had constructed for this ontology in previous work \cite{Androutsopoulos2013}. We added more sentence plans to ensure that three sentence plans were available per relation.\footnote{The domain-dependent generation resources of \nlowl that we used in all the experiments of this article are available upon request and will be made publicly available when this article is published.} A single \nl name was available per individual and class in these experiments. We generated English texts for the 52 wine individuals of the ontology; we did not experiment with texts describing classes, because we could not think of multiple alternative sentence plans for many of their axioms. For each wine individual, there were 5 available facts on average and a maximum of 6 facts. 

We generated texts with \ilpnlg, \pipeline, and \pipelineshort for the 52 individuals. With \pipeline and \pipelineshort, we generated texts for $M = 2, 3, 4, 5, 6$; recall that $M$ is the number of selected facts per text, and that for each $M$ value the texts of \pipeline and \pipelineshort are generated three times, with randomly selected sentence plans (Section~\ref{systemsOverview}). With \ilpnlg, we repeated the generation of the texts of the 52 individuals using different $\lambda_1$ values ($\lambda_2 = 1 - \lambda_1$), which led to texts expressing from zero to all of the available facts. We set the maximum number of fact subsets to $m = 3$, which was the maximum number of sentences (after aggregation) in the texts of \pipeline and \pipelineshort. All three systems were allowed to form aggregated sentences with up to $B_{max} = 22$ distinct elements; this was the number of distinct elements of the longest aggregated sentence in our previous experiments \cite{Androutsopoulos2013}, where \pipeline was allowed to 
combine up to three simple (expressing one fact each) sentences to form an aggregated one.\footnote{We modified \pipeline and \pipelineshort to count distinct elements during aggregation.} 

\begin{figure}
\center
\includegraphics[width=0.7\columnwidth]{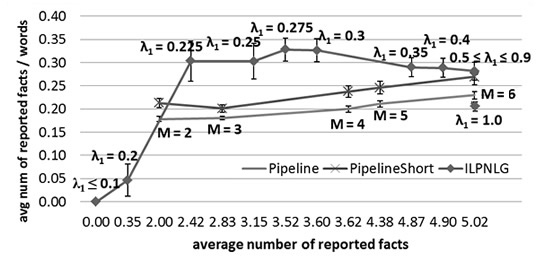}
\caption{Facts per word ratios for the Wine Ontology (grouped by $M$ or $\lambda_1$ values).}
\label{ILPResultsWineFW}
\end{figure}

For each $M$ value (in the case of \pipeline and \pipelineshort) and for each $\lambda_1$ value (in the case of \ilpnlg), we measured the average (over the 52 texts) number of facts each system reported per text (horizontal axis of Fig.~\ref{ILPResultsWineFW}), and the average (again over the 52 texts) number of facts each system reported per text divided by the average (over the 52 texts) number of words (vertical axis of Fig.~\ref{ILPResultsWineFW}, with error bars showing 95\% confidence 
intervals).\footnote{For $0.5 \leq \lambda_1 \leq 0.9$, \ilpnlg's results were identical.} As one would expect, \pipelineshort expressed on average more facts per word (Fig.~\ref{ILPResultsWineFW})
than \pipeline, but the differences were small. 

For 
$\lambda_1 \leq 0.1$
(far left of Fig.~\ref{ILPResultsWineFW}), \ilpnlg produces empty texts, because it focuses on minimizing the number of distinct elements of each text. For $\lambda_1 \geq 0.2$, it performs better than \pipeline and \pipelineshort. For $\lambda_1 \approx 0.3$, it obtains the highest average facts per word ratio by selecting the facts and sentence plans that lead to the most compressive aggregations. For greater values of $\lambda_1$, it selects additional facts whose sentence plans do not aggregate that well, which is why the ratio declines. When $M$ is small, the two pipeline systems often select facts and sentence plans that offer few aggregation opportunities; as the number of selected facts increases, some more aggregation opportunities arise, which is why the facts per word ratio 
of the two systems improves. 

\begin{figure}
\center
\includegraphics[width=0.7\columnwidth]{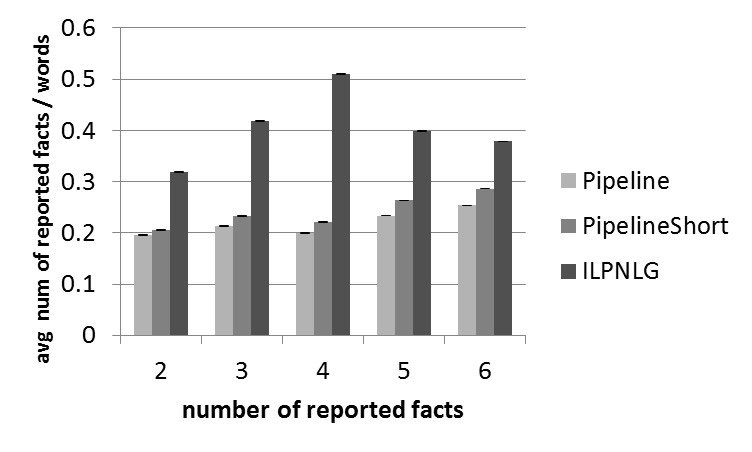}
\caption{Facts per word ratios for the Wine Ontology (grouped by numer of reported facts).}
\label{ILPResultsWineFWBars}
\end{figure}

Figure~\ref{ILPResultsWineFWBars} provides an alternative view of the behavior of the three systems. In this case, we group together all the texts of each system (regardless of the $M$ or $\lambda_1$ values that were used to generate them) that report 2, 3, 4, 5, or 6 facts (horizontal axis of Fig.~\ref{ILPResultsWineFWBars}). For each group (and each system), we show (vertical axis of Fig.~\ref{ILPResultsWineFWBars}) the 
average number of reported facts per text, divided by the average number of words of the texts in the group.\footnote{We remove from each group duplicate texts of the same system (for different $M$ or $\lambda_1$ values). If we still have more than one texts (of the same system) for the same individual or class in the same group, we keep only the text with the best facts per word ratio, to avoid placing too much emphasis on individuals and classes with many texts in the same group. Especially for \pipeline, whose texts are generated three times per individual and class (for each $M$ value), we keep the three texts (regardless of $M$ value, excluding duplicates) with the highest facts per word ratios per individual or class in each group.} Again, Fig.~\ref{ILPResultsWineFWBars} shows that \ilpnlg produces clearly more compact texts than \pipeline and \pipelineshort, with the difference between the latter two systems being very small.\footnote{Figure~\ref{ILPResultsWineFWBars} and all the similar figures in the remainder of this article include error bars corresponding to 95\% confidence intervals, but the intervals 
are so small that they can hardly be seen.} In all the experiments of this section, the \ilp solver (used in \ilpnlg) was very fast (average: 0.08 sec, worst: 0.14 sec per text). 

We show below sample texts generated by \pipeline and \pipelineshort (both with $M = 4$) and \ilpnlg (with $\lambda_1 = 0.3$). 

\bigskip
{\small
\noindent
\pipeline: This Sauternes has strong flavor. It is made from 
Sauvignon Blanc and Semillon grapes and it is produced by Chateau D'ychem.\\
\pipelineshort: This is a strong Sauternes. It is made from Sauvignon Blanc and Semillon grapes and it is produced by Chateau D'ychem.\\
\ilpnlg: This is a strong Sauternes. It is made from Sauvignon Blanc and Semillon grapes by Chateau D'ychem.
}

\bigskip

{\small
\noindent 
\pipeline: This Riesling has sweet taste and it is full bodied. It is made by Schloss Volrad.\\
\pipelineshort: This is a full sweet Riesling. It is produced by Schloss Volrad. \\
\ilpnlg: This is a full sweet moderate Riesling.
}
\bigskip

\noindent In the first group of generated texts above, \pipeline and \pipelineshort use different verbs for the grapes and producer, whereas \ilpnlg uses the same verb, which leads to a more compressive aggregation; all the texts of the first group describe the same wine and report four facts each. In the second group of generated texts above, \ilpnlg has chosen to report the (moderate) flavor of the wine instead of the producer, and uses the same verb (`is') for all the facts, leading to a shorter sentence; again all the texts of the second group describe the same wine and report four facts each. Recall that we treat all (non-redundant) facts as equally important in our experiments. In both groups of texts, some facts are not aggregated because they belong in different topical sections.

We also wanted to investigate the effect of the higher facts per word ratio of \ilpnlg on the perceived quality of the generated texts, compared to the texts of the pipeline systems. We were concerned that the more compressive aggregations of \ilpnlg might lead to sentences sounding less fluent or unnatural, though aggregation is often used to  produce more fluent texts. We were also concerned that the more compact texts of \ilpnlg might be perceived as being more difficult to understand (less clear) or less well-structured. To investigate these issues, we showed the $52 \times 2 = 104$ texts of \pipelineshort ($M=4$) and \ilpnlg ($\lambda_1 = 0.3$) to 6 computer science students (undergraduates and graduates), who were not involved in the work of this article; they were all fluent, though not native English speakers. We did not use \pipeline in this experiment, since its facts per word ratio was similar to that of \pipelineshort.  Each one of the 104 texts was given to exactly one student. Each student was given approximately 9 randomly selected texts of each system. The \owl statements that the texts were generated from were not shown, and the students did not know which system had generated each text. Each student was shown all of his/her texts in random order, regardless of the system that generated them. The students were asked to score each text by stating how strongly they agreed or disagreed with statements $S_1$--$S_3$ below. A scale from 1 to 5 was used (1: strong disagreement, 3: ambivalent, 5: strong agreement). 

\medskip
{\small
($S_1$) \emph{Sentence fluency}: The sentences of the text are fluent, i.e., each sentence on its own is grammatical and sounds natural. When two or more smaller sentences are combined to form a single, longer sentence, the resulting longer sentence is also grammatical and sounds natural.

($S_2$) \emph{Text structure}: The order of the sentences is appropriate. The text presents information by moving reasonably from one topic to another. 

($S_3$) \emph{Clarity}: The text is easy to understand, if the reader is familiar with basic wine terms. 
}
\medskip

\noindent The students were also asked to provide an overall score (1--5) per text. We did not score referring expressions, since both systems use the same component for them. We note that although both systems use the same text planner, in \pipelineshort (and all the pipeline variants) the text planner is invoked once, whereas in \ilpnlg (and \ilpnlgextend) it is invoked at different stages before and after using the \ilp model (Section~\ref{systemsOverview}), which is why we collected text structure scores too.

\begin{table}
\caption{Human scores for Wine Ontology texts.}
\label{wine_results}
{\small
\begin{tabular}{lcc}
Criteria              & \pipelineshort    & \ilpnlg \\
\hline 
Sentence fluency      & $4.75 \,\pm 0.21$ & $\textbf{4.85} \,\pm 0.10$ \\
Text structure        & $\textbf{4.94} \,\pm 0.06$ & $4.88 \,\pm 0.14$ \\
Clarity               & $\textbf{4.77} \,\pm 0.18$ & $4.75 \,\pm 0.15$ \\
Overall               & $4.52 \,\pm 0.20$ & $\textbf{4.60} \,\pm 0.18$ \\
\hline 
\end{tabular}
}
\end{table}

Table \ref{wine_results} shows the average scores of the two systems with 95\% confidence intervals.
For each criterion, the best score is shown in bold. The sentence fluency and overall scores of \ilpnlg are slightly higher than those of \pipelineshort, whereas \pipelineshort obtained a slightly higher score for text structure and clarity. The differences, however, are very small, especially in clarity, and we detected no statistically significant difference between the two systems in any of the criteria.\footnote{We performed Analysis of Variance (\textsc{anova}) and post-hoc Tukey tests to check for statistically significant differences. A post-hoc power analysis of the \textsc{anova} values resulted in power values greater or equal to $0.95$. We also note that in similar previous experiments \cite{Androutsopoulos2013}, inter-annotator agreement was strong (sample Pearson correlation $r \geq 0.91$).} Hence, there was no evidence in these experiments that the higher facts per word ratio of \ilpnlg comes at the expense of lower perceived text quality. We investigated these issues further in a second set of experiments, discussed in the next section, where the generated texts were longer.

\subsection{Experiments with the Consumer Electronics Ontology} \label{productExperiments}

\begin{figure}
\begin{minipage}[c]{.5\columnwidth}
\center
\includegraphics[width=\columnwidth]{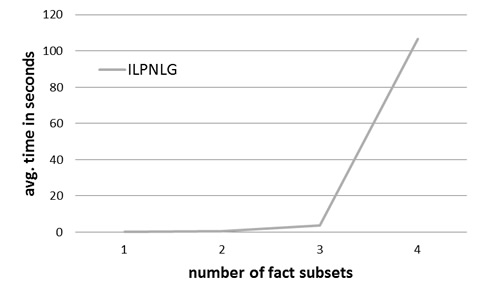}
\caption{Average solver times for \ilpnlg with different maximum numbers of fact subsets ($m$),
for the Consumer Electronics ontology.}
\label{ILPTimeSubset}
\end{minipage}
\begin{minipage}[c]{.5\columnwidth}
\center
\includegraphics[width=\columnwidth]{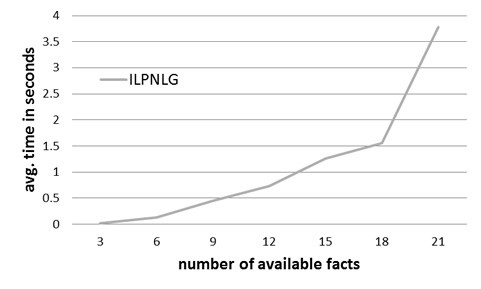}
\caption{Average solver times for \ilpnlg with different numbers of available facts ($|F|$) and $m=3$, 
for the Consumer Electronics ontology.}
\label{ILPTimeFact}
\end{minipage}
\end{figure}

In the second set of experiments, we used the Consumer Electronics Ontology, with the manually authored domain-dependent generation resources (e.g., text plans, \nl names, sentence plans) of our previous work \cite{Androutsopoulos2013}. As in the previous section, we added more sentence plans to ensure that three sentence plans were available for almost every relation; for some relations we could not think of enough sentence plans. Again, a single \nl name was available per individual and class. 

\begin{figure}
\begin{minipage}[c]{.5\columnwidth}
\center
\includegraphics[width=\columnwidth]{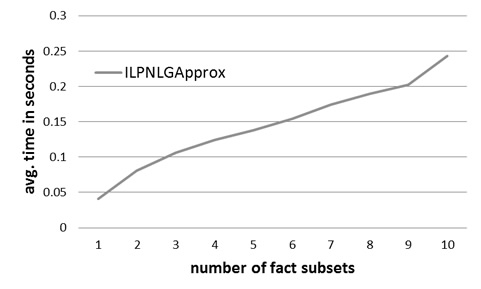}
\caption{Average solver times for \ilpnlgapprox with different numbers of fact subsets ($m$), 
for the Consumer Electronics ontology.}
\label{ILPApproxTimeSubset}
\end{minipage}
\begin{minipage}[c]{.5\columnwidth}
\center
\includegraphics[width=\columnwidth]{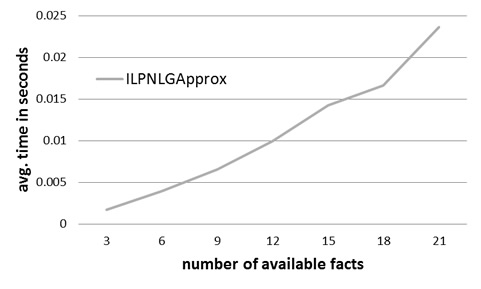}
\caption{Average solver times for \ilpnlgapprox with different numbers of available facts ($|F|$) and $m=3$, 
for the Consumer Electronics ontology.}
\label{ILPApproxTimeFact}
\end{minipage}
\end{figure}

We generated English texts with \ilpnlg, \pipeline, \pipelineshort for the 30 development individuals (Section~\ref{ontologiesOfExperiments}), using $M = 3, 6, 9, \dots, 21$ in the two pipeline systems, and different values of $\lambda_1$ ($\lambda_2 = 1 - \lambda_1$) in \ilpnlg. All three systems were allowed to form aggregated sentences with up to $B_{max} = 39$ distinct elements; this was the number of distinct elements of the longest aggregated sentence in the experiments of our previous work \cite{Androutsopoulos2013}, where \pipeline was allowed to combine up to three simple (expressing one fact each) sentences to form an aggregated one. There are 14 available facts ($|F|$) on average and a maximum of 21 facts for each one of the 30 development individuals, compared to the 5 available facts on average and the maximum of 6 facts of the Wine Ontology. Hence, the texts of the Consumer Electronics Ontology are much longer, when they report all the available facts. In \ilpnlg, we would have to set the maximum number of fact subsets to $m = 10$, which was the maximum number of (aggregated) sentences in the texts of \pipeline and \pipelineshort. The number of variables of our \ilp model, however, grows exponentially to $m$ and $|F|$ (Fig.~\ref{ILPTimeSubset}--\ref{ILPTimeFact}), though the effect of $|F|$ is weaker.

Figure~\ref{ILPTimeSubset} shows the average time the \ilp solver took for different values of $m$ in the experiments with the Consumer Electronics ontology; the results are averaged over the 30 development individuals and also for $\lambda_1 = 0.4, 0.5, 0.6$. For $m=4$, the solver took 1 minute and 47 seconds on average per text; recall that $|F|$ is also much larger now, compared to the experiments of the previous section. For $m=5$, the solver was so slow that we aborted the experiment. Figure~\ref{ILPTimeFact} shows the average solver times for different numbers of available facts $|F|$, for $m = 3$; in this case, we modified the set of available facts ($F$) of every individual to contain $3, 6, 9, 12, 15, 18, 21$ facts. The results are again averaged over the 30 development individuals and for $\lambda_1 = 0.4, 0.5, 0.6$. Although the times of Fig.~\ref{ILPTimeFact} also grow exponentially to $|F|$, they remain under 4 seconds, showing that the main factor to the complexity of \ilpnlg is $m$, the number of fact subsets, i.e., the maximum allowed number of (aggregated) sentences of each text. 

To efficiently generate texts with larger $m$ values, we developed \ilpnlgapprox, the approximation of \ilpnlg that considers each fact subset separately (Section~\ref{Complexity}). Figures~\ref{ILPApproxTimeSubset}--\ref{ILPApproxTimeFact} show the average solver times of \ilpnlgapprox for different values of $m$ and $|F|$, respectively; all the other settings are as in Fig.~\ref{ILPTimeSubset}--\ref{ILPTimeFact}. The solver times now grow approximately linearly to $m$ and $|F|$ and are under 0.3 seconds in all cases.

\begin{figure}
\center
\includegraphics[width=0.8\columnwidth]{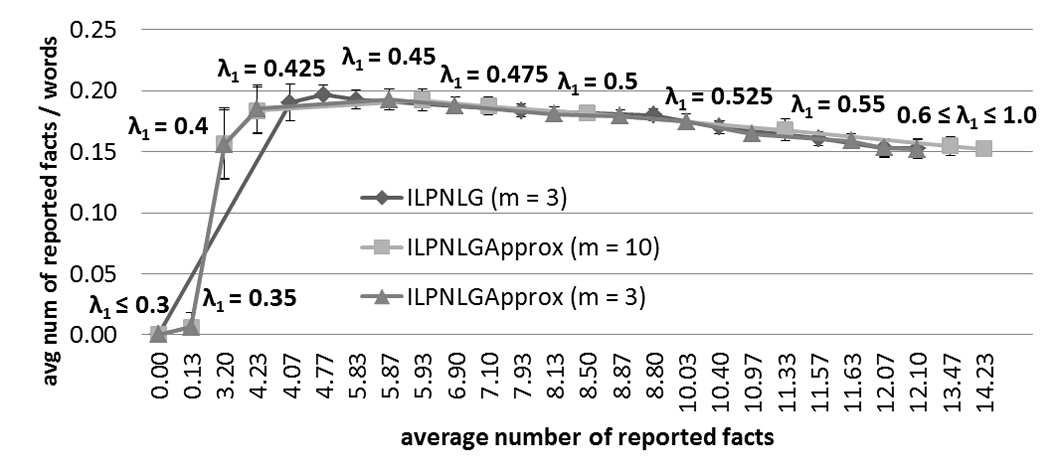}
\caption{Comparing the facts per word ratios of \ilpnlgapprox and \ilpnlg in texts generated from the Consumer Electronics ontology.}
\label{ILPResultsApproxProductFW}
\end{figure}

In Figure~\ref{ILPResultsApproxProductFW}, we compare \ilpnlg to \ilpnlgapprox, by showing their average fact per word ratios, computed as in Fig.~\ref{ILPResultsWineFW} (Section~\ref{wineExperiments}).  We set $m = 3$ in \ilpnlg to keep the solving times low; in \ilpnlgapprox we experimented with both $m = 3$ (the value used in \ilpnlg) and $m = 10$ (the value that was actually needed). In all cases, $B_{max} = 39$. The facts per word ratios of all three systems are very similar. We conclude that \ilpnlgapprox achieves very similar results to \ilpnlg in much less time.

\begin{figure}
\center
\includegraphics[width=0.8\columnwidth]{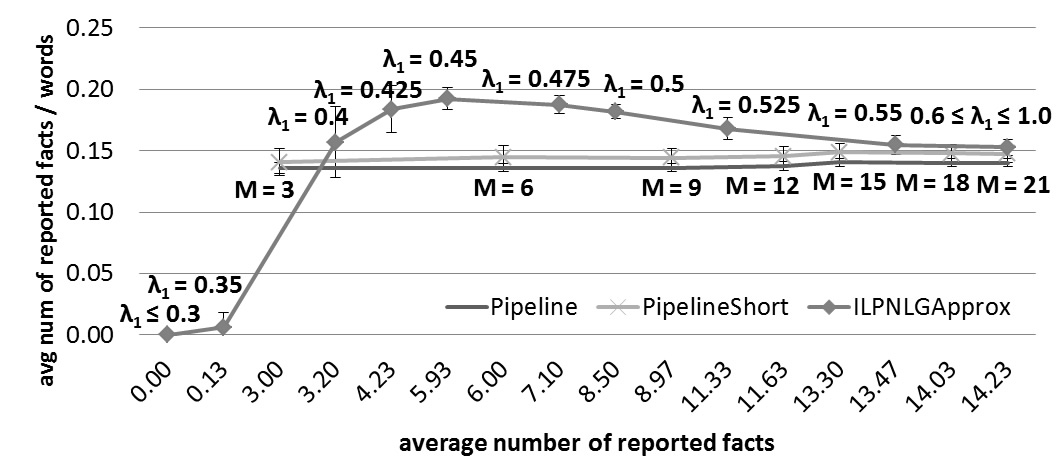}
\caption{Facts per word ratios for the Consumer Electronics Ontology (grouped by $M$ or $\lambda_1$ values).}
\label{ILPResultsProductFW}
\end{figure}

\begin{figure}
\center
\includegraphics[width=0.7\columnwidth]{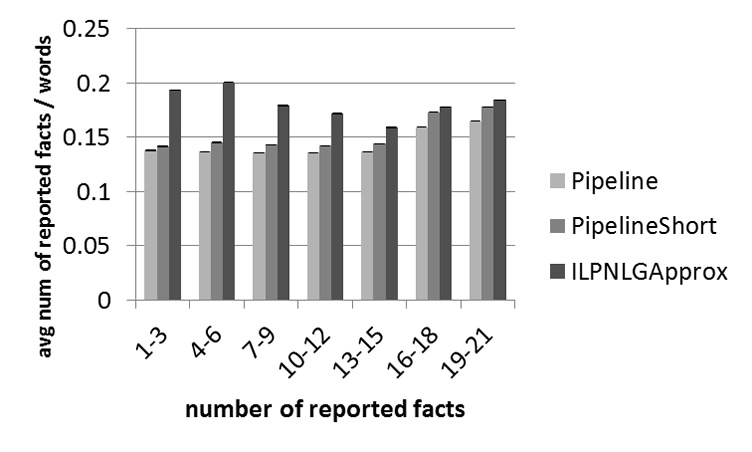}
\caption{Facts per word ratios for the Consumer Electronics Ontology (grouped by reported facts).}
\label{ILPResultsProductFWBars}
\end{figure}

Figures~\ref{ILPResultsProductFW} and \ref{ILPResultsProductFWBars} show the facts per word ratios of \ilpnlgapprox ($m = 10$), \pipeline, and \pipelineshort, computed in two ways, as in Section~\ref{wineExperiments}, for the texts of the 30 development individuals.
Again, \pipelineshort achieves slightly better results than \pipeline. The behavior of \ilpnlgapprox in Figure~\ref{ILPResultsProductFW} is very similar to the behavior of \ilpnlg on the Wine Ontology (Fig.~\ref{ILPResultsWineFW}); for 
$\lambda_1 \leq 0.3$ it produces empty texts, while for $\lambda_1 \geq 0.4$ it performs better than the other systems. \ilpnlgapprox obtains the highest facts per word ratio for $\lambda_1 = 0.45$, where it selects the facts and sentence plans that lead to the most compressive aggregations. For greater values of $\lambda_1$, it selects additional facts whose sentence plans do not aggregate that well, which is why the ratio declines. The two pipeline systems select facts and sentence plans that offer very few aggregation opportunities; as the number of selected facts increases, some more aggregation opportunities arise, which is why the facts per word ratio of the two systems improves (more clearly in Fig.~\ref{ILPResultsProductFWBars}). Figure~\ref{ILPResultsProductFWBars} also shows that \ilpnlgapprox generates more compact texts than \pipeline and \pipelineshort.

We show below three example texts produced by \pipeline, \pipelineshort (both with $M = 6$), and \ilpnlgapprox ($\lambda_1 = 0.45$, $m=10$). Each text reports six facts, but \ilpnlgapprox has selected facts and sentence plans that allow more compressive aggregations. Recall that we treat all the facts as equally important. If importance scores are also available (e.g., if dimensions are less important), they can be added as multipliers $\mathit{imp}(f_i)$ of $\alpha_i$ in the objective function (Eq.~\ref{AUEB_NLG_ILP_CS_AGG}) of the \ilp model.

\medskip
{\small
\noindent
\pipeline: SonySony DCR-TRV270 requires minimum illumination of 4.0 lux and its display is 2.5 in. It features a Sports scene mode, it includes a microphone and an IR remote control. Its weight is 780.0 grm.\\
\pipelineshort: Sony DCR-TRV270 requires minimum illumination of 4.0 lux and its display is 2.5 in. It features a Sports scene mode, it includes a microphone and an IR remote control. It weighs 780.0 grm. \\
\ilpnlgapprox: Sony DCR-TRV270 has a microphone and an IR remote control. It is 98.0 mm high, 85.0 mm wide, 151.0 mm deep and it weighs 780.0 grm.
}
\medskip

We showed the $30 \times 2 = 60$ texts of \pipelineshort ($M=6$) and \ilpnlgapprox ($\lambda_1 = 0.45$, $m=10$) to the same six students that participated in the experiments with the Wine Ontology (Section~\ref{wineExperiments}). Again, each text was given to exactly one student. Each student was given approximately 5 randomly selected texts of each system. The \owl statements were not shown, and the students did not know which system had generated each text. Each student was shown all of his/her texts in random order, regardless of the system that  generated them. The students were asked to score each text by stating how strongly they agreed or disagreed with statements $S_1$--$S_3$, as in Section~\ref{wineExperiments}. They were also asked to provide an overall score (1--5) per text. 

Table \ref{product_results} shows the average scores of the two systems with 95\% confidence intervals. 
For each criterion, the best score is shown in bold; the confidence interval of the best score is also shown in bold, if it does not overlap with the  confidence interval of the other system. Unlike the Wine Ontology experiments (Table~\ref{wine_results}), the scores of our \ilp approach (with the approximation of \ilpnlgapprox) are now higher than those of \pipelineshort in all of the criteria, and the differences are also larger, though we found the differences to be statistically significant only for clarity and overall quality.\footnote{When two confidence intervals do not overlap, the difference is statistically significant. When they overlap, the difference may still be statistically significant; we performed Analysis of Variance (\textsc{anova}) and post-hoc Tukey tests to check for statistically significant differences in those cases. A post-hoc power analysis of the \textsc{anova} values resulted in power values greater or equal to $0.95$.
} We attribute these larger differences, compared to the Wine Ontology experiments, to the fact that the texts are now longer and the sentence plans more varied, which often makes the texts of \pipelineshort sound verbose and, hence, more difficult to follow, compared to the more compact texts of \ilpnlgapprox, which sound more concise. 

\begin{table}
\caption{Human scores for Consumer Electronics texts.}
\label{product_results}
{\small 
\begin{tabular}{lcc}
Criteria              & \pipelineshort    & \ilpnlgapprox \\
\hline 
Sentence fluency      & $4.50 \,\pm 0.30$ & $\mathbf{4.87} \,\pm 0.12$ \\
Text structure        & $4.33 \,\pm 0.36$ & $\mathbf{4.73} \,\pm 0.22$ \\
Clarity               & $4.53 \,\pm 0.29$ & $\mathbf{4.97 \,\pm 0.06}$ \\
Overall               & $4.10 \,\pm 0.31$ & $\mathbf{4.73 \,\pm 0.16}$ \\
\hline
\end{tabular}
}
\end{table}

Overall, the human scores of the experiments with the Wine and Consumer Electronics ontologies suggest that the higher 
facts per word ratios of our \ilp approach do not come at the expense of lower perceived text quality. On the contrary, the texts of the \ilp approach may be perceived as clearer and overall better than those of the pipeline, when the texts report many facts.

\subsection{Experiments with the Disease Ontology} \label{diseaseExperiments}

In a third set of experiments, we generated texts for the 200 `development' classes (Section~\ref{ontologiesOfExperiments}) of the Disease Ontology, using the manually authored domain-dependent generation resources (e.g., text plans, \nl names, sentence plans) of Evaggelakaki \shortcite{Evaggelakaki2014}, but with additional sentence plans we constructed to ensure that there were three alternative sentence plans per relation. We generated texts with \ilpnlg, \pipeline, and \pipelineshort, for $M = 2, 3, 4, \dots, 7$ in the two pipeline systems, and different values of $\lambda_1$ ($\lambda_2 = 1 - \lambda_1$) in \ilpnlg. All three systems were allowed to form aggregated sentences with up to $B_{max} = 30$ distinct elements; this was the number of distinct elements of the longest aggregated sentence in the experiments of Evaggelakaki \shortcite{Evaggelakaki2014}, where \pipeline was allowed to combine up to three simple (expressing one fact each) sentences to form an aggregated one. There are 3.7 available facts ($|F|$) on average and a maximum of 7 facts for each one of the 200 classes. In \ilpnlg, we set $m = 4$, which was the maximum number of (aggregated) sentences in the texts of \pipeline and \pipelineshort. We did not use \ilpnlgapprox, since \ilpnlg was reasonably fast (average solver time: 0.11 sec per text, worst: 0.90 sec per text), because of the smaller values of $m$ and $|F|$, compared to the experiments of the Consumer Electronics ontology.

\begin{figure}
\center
\includegraphics[width=0.8\columnwidth]{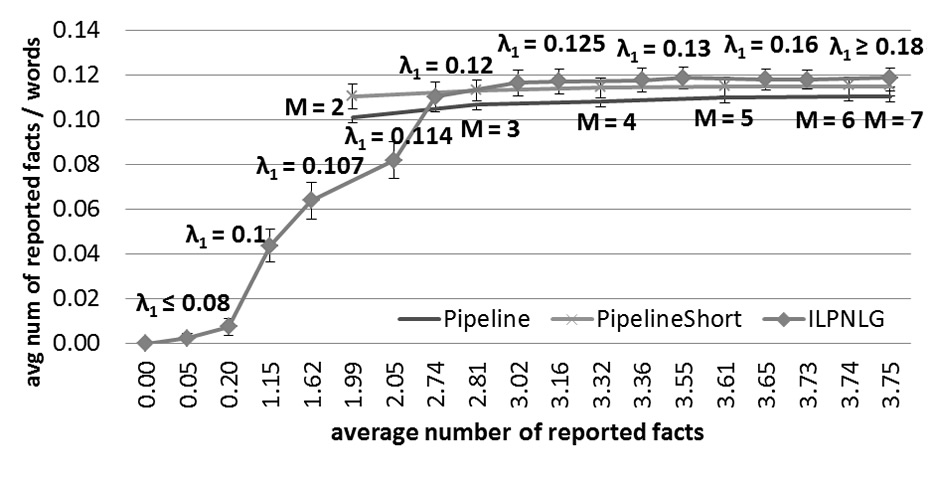}
\caption{Facts per word ratios (grouped by $M$ or $\lambda_1$ values) of \ilpnlg, \pipeline, and \pipelineshort for texts generated from the Disease Ontology.}
\label{ILPResultsDiseaseFW}
\end{figure}

\begin{figure}
\center
\includegraphics[width=0.7\columnwidth]{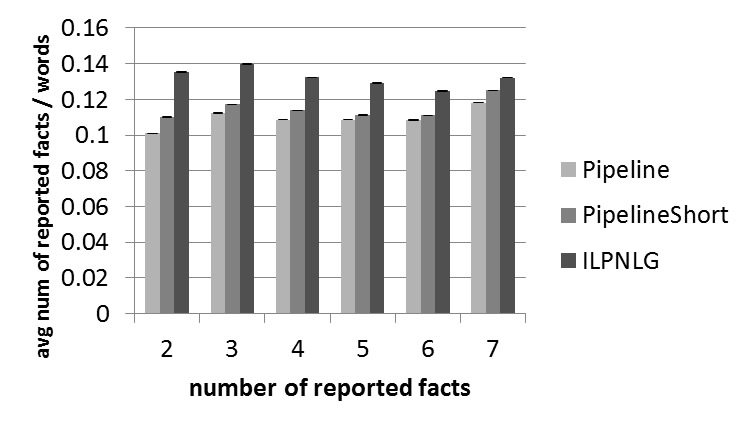}
\caption{Facts per word ratios (grouped by reported facts) of \ilpnlg, \pipeline, and \pipelineshort for texts generated from the Disease Ontology.}
\label{ILPResultsDiseaseFWBars}
\end{figure}

Figures~\ref{ILPResultsDiseaseFW} and \ref{ILPResultsDiseaseFWBars} show the facts per word ratios of \ilpnlg, \pipeline, and \pipelineshort, computed in two ways, as in Section~\ref{wineExperiments}, for the texts of the 200 classes. \pipelineshort achieves only slightly better results than \pipeline in both figures. Also, Fig.~\ref{ILPResultsDiseaseFWBars} shows that \ilpnlg produces more compact texts than the two pipeline systems. In Figure~\ref{ILPResultsDiseaseFW}, however, the difference between \ilpnlg and the two pipeline systems is less clear. For small $\lambda_1$ values, \ilpnlg produces empty texts, because it focuses on minimizing the number of distinct elements of each text. For $\lambda_1 \geq 0.125$, it performs only marginally better than \pipelineshort, unlike previous experiments (cf.~Fig.~\ref{ILPResultsWineFW} and \ref{ILPResultsProductFW}). We attribute this difference to the fact that \ilpnlg does not take into account the lengths of the \nl names, which vary a lot in the Disease Ontology; nor does it take into account that the $O$ of many facts $\left<S,R,O\right>$ is a conjunction. These issues were addressed in our extended \ilp model (Section~\ref{OurExtendedModel}), which is used in \ilpnlgextend.

\begin{figure}
\center
\includegraphics[width=0.8\columnwidth]{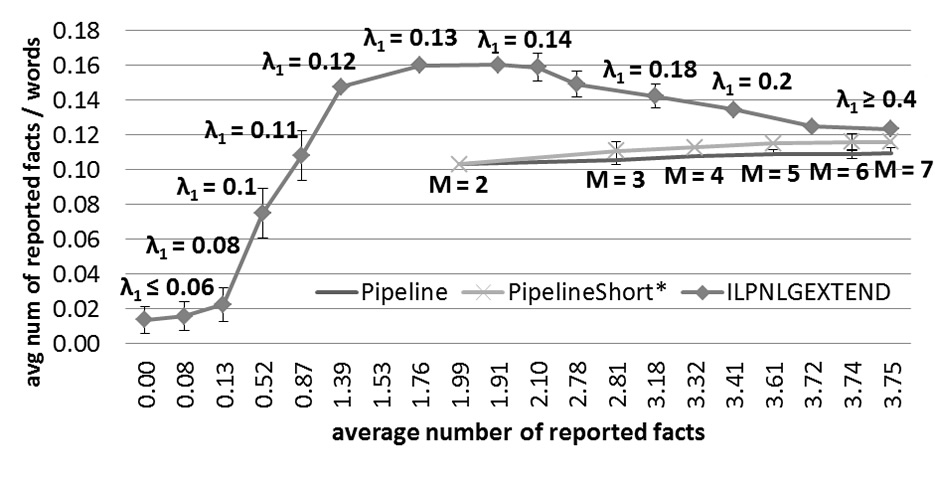}
\caption{Facts per word ratios (grouped by $M$ or $\lambda_1$ values) of \ilpnlgextend, \pipeline, and \pipelineshortnln for texts generated from the Disease Ontology.}
\label{ILPResultsDiseaseFWExtend}
\end{figure}

\begin{figure}
\center
\includegraphics[width=0.7\columnwidth]{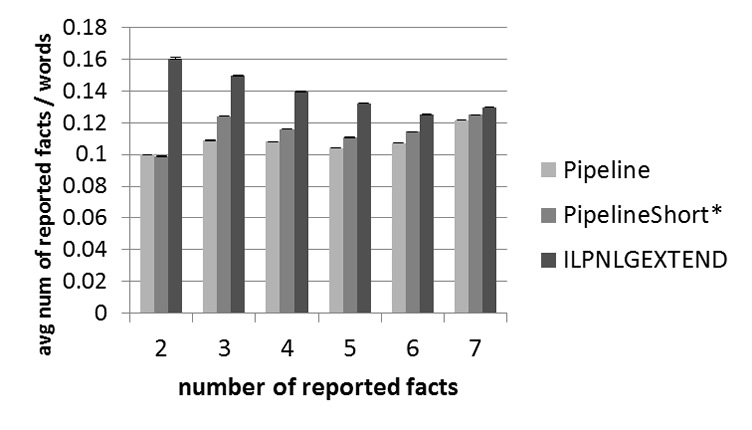}
\caption{Facts per word ratios (grouped by reported facts) of \ilpnlgextend, \pipeline, and \pipelineshortnln for texts generated from the Disease Ontology.}
\label{ILPResultsDiseaseFWExtendBars}
\end{figure}

We then generated texts for the 200 classes again, this time with \pipeline, \pipelineshortnln (both with $M = 2, 3, \dots,  7$, $W_{max} = 54$) and \ilpnlgextend ($m$ = 4, $W_{max} = 54$); we modified \pipeline and \pipelineshort to count 
words (instead of elements) when comparing against \ilpnlgextend, which is why we report $W_{max}$ in all three systems. Similarly to how $B_{max}$ was previously selected, $W_{max}=54$ was the number of words of the longest aggregated sentence in the experiments of Evaggelakaki \shortcite{Evaggelakaki2014}. Figures~\ref{ILPResultsDiseaseFWExtend} and \ref{ILPResultsDiseaseFWExtendBars} show the new facts per word ratios, for the texts of the 200 classes. In Figure~\ref{ILPResultsDiseaseFWExtend}, for $\lambda_1 \leq 0.06$, \ilpnlgextend produces empty texts, because it focuses on minimizing the lengths of the texts. For $\lambda_1 \geq 0.12$, \ilpnlgextend now performs clearly better than the pipeline systems, obtaining the highest facts per word ratio for $\lambda_1 = 0.14$; notice that we now compare to \pipelineshortnln, which is a better baseline for \ilpnlgextend than \pipelineshort (Section~\ref{systemsOverview}). Figure~\ref{ILPResultsDiseaseFWExtendBars} also confirms that \ilpnlgextend outperforms the pipeline systems. 
The \ilp solver was actually slightly faster with \ilpnlgextend (average: 0.09 sec, worst: 0.65 sec per text) compared to \ilpnlg (average: 0.11 sec, worst: 0.90 sec per text).

We show below three example texts produced by \pipeline, \pipelineshortnln (both with $M = 3$), and \ilpnlgextend ($\lambda_1 = 0.14$). 
Each text reports three facts, but \ilpnlgextend has selected 
facts with fewer and shorter \nl names, and sentence plans that lead to better sentence aggregation. 
Recall that we treat all facts 
as equally important in these experiments, but that our \ilp models can also handle importance scores (e.g., treating facts reporting symptoms as more important than facts about is-a relations).

\vspace*{3mm}
{\small
\noindent 
\pipeline: Nephropathia epidemica can be found in the kidneys. It can often cause myalgia, nausea, renal failure, vomiting, abdominal pain, headaches, internal hemorrhage and back pain, and it results in infections. \\
\pipelineshortnln: Nephropathia epidemica is a kind of hemorrhagic fever with renal syndrome. It originates from bank voles and it is caused by the puumala virus. \\
\ilpnlgextend: Nephropathia epidemica results in infections. It often originates from bank voles from the puumala virus. 
}

\subsection{Further Experiments with the Wine Ontology} \label{wineMoreExperiments}

The experiments of the previous section tested the ability of \ilpnlgextend to take into account the different lengths of \nl names and the fact that some facts $\left<S,R,O\right>$ involve conjunctions (or disjunctions) in their $O$. They did not, however, test the ability of \ilpnlgextend to cope with multiple alternative \nl names per individual or class. The Consumer Electronics and Disease Ontologies were inappropriate in this respect, because the names of electronic products tend to be unique and we did not have the expertise to create alternative names of diseases, as already noted. Instead, we returned to the Wine Ontology, which had been used in Section~\ref{wineExperiments} with a single \nl name per individual and class. We now added more \nl names to the Wine Ontology to ensure that approximately three \nl names on average (with a minimum of 2 and a maximum of 5) were available for each one of the individual and classes we generated texts for. We generated texts for the 52 wine individuals and 24 of the wine classes of the Wine Ontology, using \pipeline, \pipelineshortnln, and \ilpnlgextend.\footnote{In the experiments of Section~\ref{wineExperiments}, we had not generated texts for wine classes, because we could not think of alternative sentence plans for their axioms. In the experiments of this section, we generated texts for 24 (out of 63) wine classes, because we were able to provide alternative \nl names for them.} 

All three systems were allowed to form aggregated sentences with up to $W_{max} = 26$ 
words; again, we modified \pipeline and \pipelineshort to count 
words (instead of elements) when comparing against \ilpnlgextend, which is why we report $W_{max}$ for all three systems. Similarly to Section~\ref{diseaseExperiments}, $W_{max}$ was set to the number of 
words of the longest aggregated sentence in the experiments of our previous work \cite{Androutsopoulos2013}, where \pipeline was allowed to combine up to three simple (expressing one fact each) sentences to form an aggregated one. In \ilpnlgextend, we used different values for $\lambda_1$ ($\lambda_2 = 1 - \lambda_1$), setting $m = 3$, as in Section~\ref{wineExperiments}. In \pipeline and \pipelineshortnln, we used  $M = 2, 3, \dots, 7$.\footnote{Generating texts for the additional 24 classes required raising the maximum $M$ value to 7, unlike the experiments of Section~\ref{wineExperiments}, where it was 6.} For each $M$ value, the texts of \pipeline  for the 76 individuals and classes were generated 10 times (not 3, unlike all the previous experiments with \pipeline); each time, we used one of the different alternative sentence plans for each relation and one of the different alternative \nl names for the individual or class the text was being generated for, since \pipeline cannot select among alternative \nl names (and sentence plans) by itself. 

\begin{figure}
\center
\includegraphics[width=0.8\columnwidth]{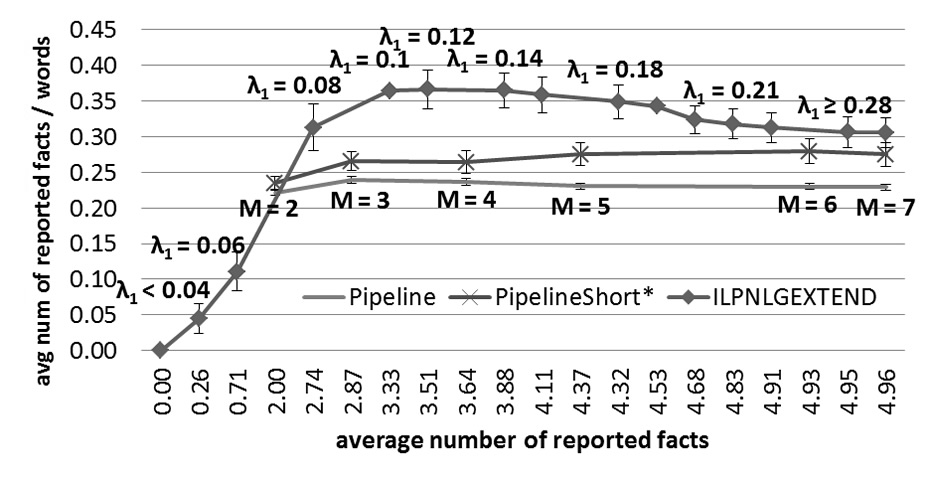}
\caption{Fact per word ratios (grouped by $M$ or $\lambda_1$ values) of the additional Wine Ontology experiments.}
\label{ILPResultsWineExtend}
\end{figure}

\begin{figure}
\center
\includegraphics[width=0.7\columnwidth]{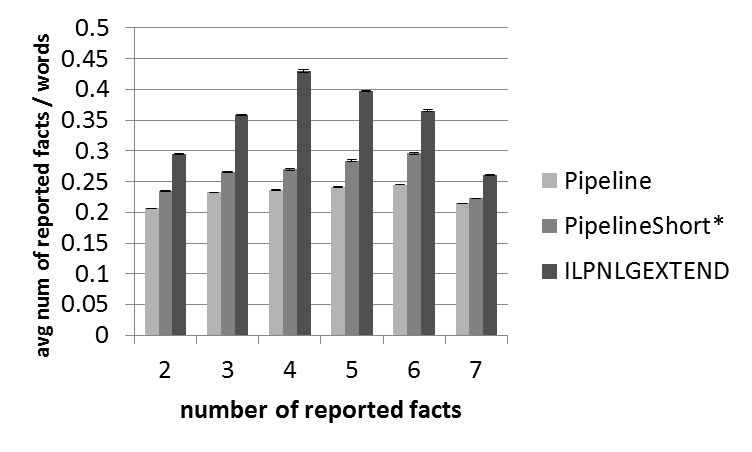}
\caption{Fact per word ratios (grouped by reported facts) of the additional Wine Ontology experiments.}
\label{ILPResultsWineExtendBars}
\end{figure}

Figures~\ref{ILPResultsWineExtend} and \ref{ILPResultsWineExtendBars} show the facts per word ratios, computed in two ways, as in Section~\ref{wineExperiments}. In Fig.~\ref{ILPResultsWineExtend}, for $\lambda_1 < 0.04$, \ilpnlgextend produces empty texts, because it focuses on minimizing the length of each text. For $\lambda_1 \geq 0.08$, it performs clearly better than the other systems. For $\lambda_1 = 0.12$, it obtains the highest facts per word ratio by selecting the facts and sentence plans that lead to the shortest (in words) aggregated sentences, and \nl names that indirectly express facts (not requiring separate sentences). 
For greater values of $\lambda_1$, \ilpnlgextend selects additional facts whose sentence plans do not aggregate that well or that cannot be indirectly expressed via \nl names, which is why the ratio of \ilpnlgextend declines. 
We note that the highest average facts per word ratio of \ilpnlgapprox (0.37, for $\lambda_1 = 0.12$) of Fig.~\ref{ILPResultsWineExtend} is higher than the highest 
average ratio (0.33, for $\lambda_1 = 0.3$) we had obtained in Section~\ref{wineExperiments} with \ilpnlg (Fig.~\ref{ILPResultsWineFW}). Also, the overall values of $\lambda_1$ are now smaller. This is due to the larger number of factors in the right part of the objective function (Eq.~\ref{AUEB_NLG_ILP_CS_AGG_EXTEND}) of \ilpnlgextend. Figure~\ref{ILPResultsWineExtendBars} confirms that \ilpnlgextend outperforms the pipelines. In the experiments of this section with \ilpnlgextend, the \ilp solver was very fast (average: 0.06 sec, worst: 0.64 sec per text). 

We show below texts produced by \pipeline, \pipelineshortnln (both with $M = 4$), and \ilpnlgextend ($\lambda_1 = 0.12$). All texts describe the same wine and report four facts.

\bigskip
{\small
\noindent 
\pipeline: This Sauvignon Blanc is dry and medium. It is made by Stonleigh and it is produced in New Zealand.\\
\pipelineshortnln: 
This delicate tasting and dry Sauvignon Blanc wine originates 
from New\\ Zealand.\\
\ilpnlgextend: 
This Stonleigh Sauvignon Blanc is dry, delicate and medium. 
}
\bigskip

\noindent \ilpnlgextend chose an \nl name that avoids expressing the maker as a separate sentence, and used the same verb (``is'') to express the other three facts, allowing a single aggregated sentence to be formed. It also avoided expressing the origin (New Zealand), which would require a long sentence that would not aggregate well with the others.

\section{Related Work} \label{RelatedWork}

Marciniak and Strube \shortcite{Marciniak2005b} proposed an \ilp approach 
to language processing problems where the decisions of classifiers that consider 
different, but co-dependent, subtasks need to be combined. They  
applied their approach to the generation of multi-sentence route directions,
by training classifiers (whose decisions affect the generated text) on a parallel corpus consisting of 
semantic representations and 
route directions. 
The classifiers control the ordering and lexicalization of 
phrases 
and a simple form of aggregation (mainly the choice of connectives between the phrases). 
Marciniak and 
Strube aimed to generate fluent and grammatically correct texts; 
by contrast, our \ilp models employ 
manually authored linguistic resources 
that guarantee fluent and grammatical texts (as also confirmed by our experiments), and make no decisions 
directly affecting fluency or grammaticality. 
Instead, our models make decisions related to content selection, lexicalization, 
aggregation (using more complex rules than Marciniak and Strube),
and a limited form of referring expression generation (in the case of our extended model), aiming to produce more compact texts, 
without invoking classifiers.

Barzilay and Lapata \shortcite{Barzilay2005} treated content selection as an optimization problem. Given a pool of facts (database entries) and scores indicating the importance of including or excluding each fact or pair of facts, their method selects the facts to express by solving an optimization problem similar to energy minimization. A solution is found by applying a minimal cut partition algorithm to a graph representing the pool of facts and the importance scores. The importance scores of single facts are obtained via supervised machine learning (AdaBoost) from a  dataset of (sports) facts and news articles expressing them. The importance scores of pairs of facts depend on parameters tuned on the same dataset using Simulated Annealing. Our \ilp models are simpler, in that they allow importance scores to be associated only with single facts, not pairs of facts. On the other hand, our models jointly perform content selection, lexicalization, aggregation, and (limited) referring expression generation, not just content selection. 

In other work, Barzilay and Lapata \shortcite{Barzilay2006} consider sentence aggregation. Given a set of facts (again database entries) that a content selection stage has produced, aggregation is viewed as the problem of partitioning the facts into optimal subsets (similar to the buckets of our \ilp models). Sentences expressing facts of the same subset are aggregated to form a longer sentence. The optimal partitioning maximizes the pairwise similarity of the facts in each subset, subject to constraints that limit the number of subsets and the number of facts in each subset. A Maximum Entropy classifier predicts the semantic similarity of each pair of facts, and an \ilp model is used to find the optimal partitioning. By contrast, our \ilp models aggregate sentences by minimizing the distinct elements of each subset, to maximize the aggregation opportunities in each subset, taking care not to aggregate together sentences expressing facts from different topics; an external text planner partitions the available facts into topical sections. Again, our models have broader scope, in the sense that they (jointly) perform content selection, lexicalization, aggregation, and (limited) referring expression generation, not just 
aggregation.

Althaus et al.\ \shortcite{Althaus2004} show that the ordering of a set of sentences to maximize local (sentence-to-sentence) coherence is equivalent to the traveling salesman problem and, hence, 
\textsc{np}-complete. They also provide an \ilp formulation of the problem, which can be solved efficiently in practice using branch-and-cut with cutting planes. Our models do not order the sentences (or facts) of the generated texts, relying on an external text planner instead. It would be particularly interesting to add sentence (or fact) ordering to our models, along the lines of Althaus et al.\ in future work.

Kuznetsova et al.\ \shortcite{Kuznetsova2012} use \ilp to generate image captions. They train classifiers 
to detect the objects in each image. Having identified the objects of a given image, they retrieve phrases from the captions of a corpus of images, focusing on the captions of objects that are similar (color, texture, shape) to the ones in the given image. To select which objects of the image to report (a kind of content selection) and in what order, Kuznetsova et al.\ maximize (via \ilp) the mean of the confidence scores of the object detection classifiers and the sum of the co-occurrence probabilities of the objects that will be reported in adjacent positions in the caption. The co-occurrence probabilities are estimated from a corpus of captions. Having decided which objects to report and their order, a second \ilp model decides which phrases to use for each object (a kind of lexicalization) and orders the phrases. The second \ilp model maximizes the confidence of the phrase retrieval algorithm and the local cohesion between subsequent phrases. Although generating image captions is very different to generating texts from ontologies, it may be possible to use ideas from the work of Kuznetsova et al.\ related to ordering objects (in our case, facts) and phrases  in future extensions of our models.  

Joint optimization \ilp models have also been used in multi-document text summarization and sentence compression \cite{McDonaldILP2007,ClarkeLapata2008,Bergkirkpatrick2011,Galanis2012,Woodsend2012}, where the input is text, not formal knowledge representations. Statistical methods to jointly perform content selection, lexicalization, and surface realization have also been proposed in \nlg \cite{Liang2009,Konstas2012,Konstas2012b}, but they are currently limited to generating single sentences from flat records, as opposed to generating multi-sentence texts from ontologies. 

To the best of our knowldge, our work is the first to consider content selection, lexicalization, sentence aggregation, and a limited form of referring expression generation as an \ilp joint optimization problem in multi-sentence concept-to-text generation. An earlier form of our work has already been published \cite{lampouras2013a,lampouras2013b}, but without the extended version of our \ilp model (Section~\ref{OurExtendedModel}), without the experiments on the Disease Ontology (Section~\ref{diseaseExperiments}), without the further experiments on the Wine Ontology (Section~\ref{wineMoreExperiments}), with facts per word ratios grouped only by $M$ and $\lambda_1$ values (without the results of Fig.~\ref{ILPResultsWineFWBars}, \ref{ILPResultsProductFWBars},  \ref{ILPResultsDiseaseFWExtendBars}, \ref{ILPResultsWineExtendBars}), and with much fewer details.

\section{Conclusions and Future Work}\label{Conclusions}

We presented an \ilp model that jointly considers decisions in content selection, lexicalization, and sentence aggregation to avoid greedy local decisions and produce more compact texts. An extended version of the \ilp model predicts more accurately the lengths of the generated texts and also performs a limited form of referring expression generation, by considering alternative \nl names and how they can indirectly express facts. We also defined an approximation of our models that generates separately each (possibly aggregated) sentence of the final text and is more efficient when longer texts are generated.
The \ilp models (and approximations) of this article were embedded in \nlowl, a state of the art publicly available \nlg system for \owl ontologies that used a pipeline architecture in its original form. Experiments with three ontologies confirmed that our models can express more facts per word, with no deterioration in the perceived quality of the generated texts or with improved perceived quality, compared to texts generated by a pipeline architecture. Our experiments also showed that our \ilp methods (or their approximations) are efficient enough to be used in practice. 

The work of this article is the first to consider content selection, lexicalization, sentence aggregation, and a limited form of referring expression generation as an \ilp joint optimization problem in multi-sentence concept-to-text generation. Previous work in \nlg employed a pipeline architecture, considered fewer and different processing stages, was concerned with generating single sentences, or had very different
inputs and goals. Our work could be extended to consider additional generation stages (e.g., text planning, or more referring expression generation decisions). It would also be interesting to combine the \ilp models with other user modeling components that would assign interest scores to message triples. Another valuable direction would be to combine \ilp models for concept-to-text generation and multi-document summarization, to produce texts summarizing both structured and unstructured information. 
\bibliographystyle{fullname}
\bibliography{lampouras_ilp_journal}
\end{document}